\documentclass{article}

\usepackage{microtype}
\usepackage{graphicx}
\usepackage{subfig}             
\usepackage{booktabs} 
\usepackage{hyperref}

\usepackage[accepted]{icml2024}

\usepackage{amsmath}
\usepackage{amssymb}
\usepackage{mathtools}
\usepackage{amsthm}

\usepackage[capitalize,noabbrev]{cleveref}
\usepackage{url}                    
\usepackage{longtable}             
\usepackage{xurl}                  
\usepackage{hyperref}              
\usepackage{listings}               

\lstdefinelanguage{json}{
    basicstyle=\ttfamily,
    showstringspaces=false,
    breaklines=true,
    frame=none,
    literate=
     *{0}{{{\color{blue}0}}}{1}
      {1}{{{\color{blue}1}}}{1}
      {2}{{{\color{blue}2}}}{1}
      {3}{{{\color{blue}3}}}{1}
      {4}{{{\color{blue}4}}}{1}
      {5}{{{\color{blue}5}}}{1}
      {6}{{{\color{blue}6}}}{1}
      {7}{{{\color{blue}7}}}{1}
      {8}{{{\color{blue}8}}}{1}
      {9}{{{\color{blue}9}}}{1}
      {:}{{{\color{red}{:}}}}{1}
      {,}{{{\color{red}{,}}}}{1}
      {\{}{{{\color{orange}{\{}}}}{1}
      {\}}{{{\color{orange}{\}}}}}{1}
      {[}{{{\color{orange}{[}}}}{1}
      {]}{{{\color{orange}{]}}}}{1},
}

\theoremstyle{plain}

\theoremstyle{definition}

\theoremstyle{remark}

\usepackage[textsize=tiny]{todonotes}

\icmltitlerunning{ComfyGI: Automatic Improvement of Image Generation Workflows}

\begin{document}

\twocolumn[
\icmltitle{ComfyGI: Automatic Improvement of Image Generation Workflows}

\begin{icmlauthorlist}
\icmlauthor{Dominik Sobania}{yyy}
\icmlauthor{Martin Briesch}{yyy}
\icmlauthor{Franz Rothlauf}{yyy}
\end{icmlauthorlist}

\icmlaffiliation{yyy}{Johannes Gutenberg University, Mainz, Germany}

\icmlcorrespondingauthor{Dominik Sobania}{dsobania@uni-mainz.de}

\icmlkeywords{Image Generation, Genetic improvement, Diffusion Models}

\vskip 0.3in
]

\printAffiliationsAndNotice{}  

\begin{abstract}
Automatic image generation is no longer just of interest to researchers, but also to practitioners. However, current models are sensitive to the settings used and automatic optimization methods often require human involvement. To bridge this gap, we introduce ComfyGI, a novel approach to automatically improve workflows for image generation without the need for human intervention driven by techniques from genetic improvement. This enables image generation with significantly higher quality in terms of the alignment with the given description and the perceived aesthetics. On the performance side, we find that overall, the images generated with an optimized workflow are about 50\% better compared to the initial workflow in terms of the median ImageReward score. These already good results are even surpassed in our human evaluation, as the participants preferred the images improved by ComfyGI in around 90\% of the cases.  
\end{abstract}

\section{Introduction}
\label{sec:introduction}

The quality of diffusion models for image generation has improved significantly in recent years \cite{ho2020denoising,dhariwal2021diffusion,rombach2022high}. A large number of models are available on platforms such as Hugging Face\footnote{\url{https://huggingface.co/models}} and can be freely used by everyone. Some of the recent models are even specialized, e.g., for the generation of digital art \cite{huang2022draw} or even photorealistic images \cite{saharia2022photorealistic}. In addition, many free tools have recently emerged that not only make automatic image generation usable for researchers and programmers, but also allow designers to experiment with image generation techniques via a user-friendly interface. One of these tools is ComfyUI\footnote{\url{https://www.comfy.org/}}, which has usually made the latest innovations in image generation available very quickly. Thanks to its modular approach, which makes it easy to link different models and other modules in a design workflow, ComfyUI is not only an interesting tool for beginners, but also for advanced users.

However, despite all the accessible tools, there are still many possible configurations in image generation, which can be further increased by using complex design workflows, which have a substantial influence on, e.g., the alignment with the given image description and the perceived aesthetics of the generated image \cite{wang2022diffusiondb}. Manually tuning all the prompts and settings can be time-consuming for the designer and often involves the user heavily in the evaluation process \cite{liu2022design}, which does little to improve the situation. So tools are needed that perform this optimization completely automatically and, just like ComfyUI, remain easily extensible so that they are not immediately made obsolete by new innovations in image generation.

\begin{figure*}[!h]
    \centering
    \includegraphics[width=\textwidth]{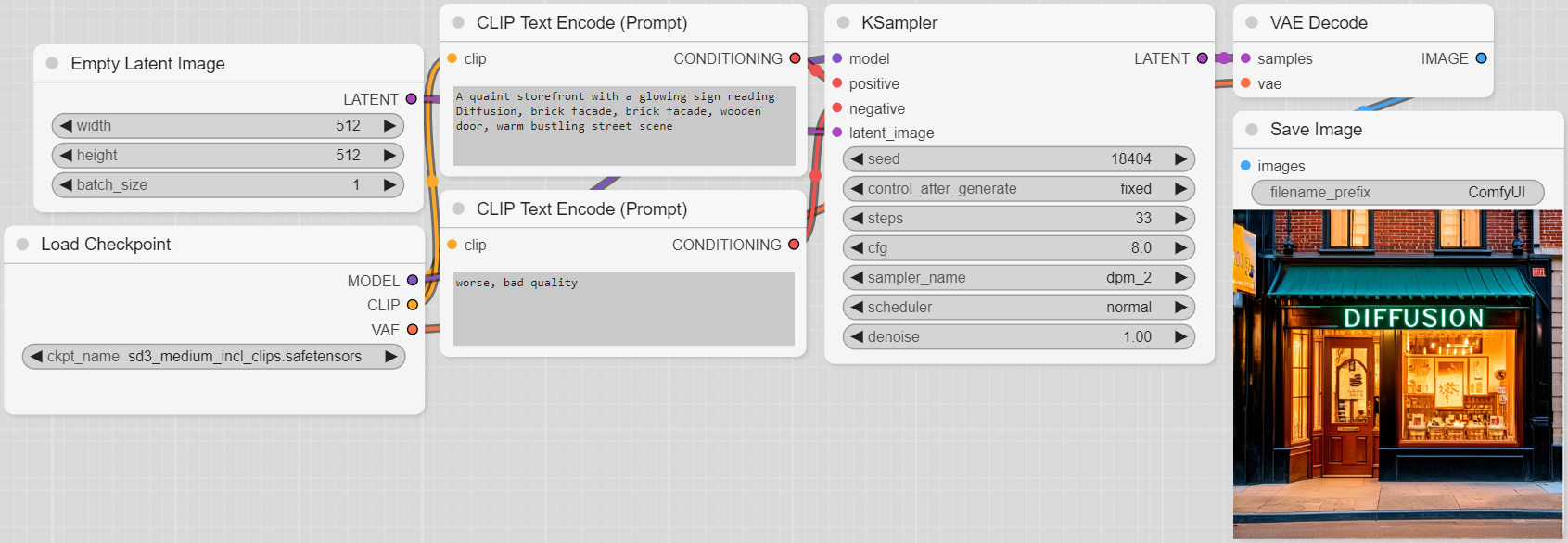}
    \caption{An example ComfyUI text-to-image workflow. The shown workflow's settings were optimized with ComfyGI and the initial prompt was \textit{``storefront with `diffusion' written on it''}.}
    \label{fig:comfyui}
\end{figure*}

In the field of software development, genetic improvement (GI) \cite{petke2017genetic} is a method that uses search-based strategies on the source code of software to optimize, for example, its non-functional properties. In software development, these are typically properties such as runtime, memory requirements, or energy consumption. This is achieved by a step-by-step improvement through small changes applied to the source code of the software (mutations) and a given objective function that guides the search process. In the image generation domain, the principles of GI can also be applied to ComfyUI's design workflows, which are processed and stored in JSON format. And with the recently released ImageReward model \cite{xu2024imagereward}, an effective objective function can be defined that evaluates generated images according to their alignment with the given description as well as their aesthetics.

Accordingly, we introduce ComfyGI\footnote{Project: \url{https://github.com/domsob/comfygi}}, the first method that applies GI techniques to ComfyUI's workflows, significantly increasing the quality of the automatically generated images in terms of the output image's alignment with the given description and its perceived aesthetics. Through its interaction with ComfyUI and simple extensibility, ComfyGI is suitable for researchers as well as practitioners. 

To improve a given design workflow and enable the generation of a high-quality image, ComfyGI uses a simple hill climbing approach. At the beginning, an image is generated using the workflow in its initial configuration and evaluated using the ImageReward model. Then, over several generations, mutations are applied to the JSON representation of the workflow and images are generated and evaluated using the modified workflow. The mutation that improves the ImageReward score the most at the end of a generation is added to the patch, which can then be applied to the JSON of the initial workflow in analogy to a software patch. The search process ends when no further improvement can be found in a generation. The mutations we use are specialized for individual components of the image generation workflow. For example, prompts can be changed using \textit{switch}, \textit{copy}, and \textit{remove} operations or by using large language models (LLMs). In addition, the sampling configuration as well as the used checkpoint model can also be changed by the mutation operators.

We analyzed ComfyGI's text-to-image generation performance on $42$ prompts out of $14$ different categories from \cite{kuimagenhub} over $10$ runs. Overall, we find that the median ImageReward score could be significantly improved by about 50\% compared to the initial images with ComfyGI. Considering the individual prompt categories (ranging from rare words and misspellings to text generation in images), an improvement can be observed for all of them. In addition, we also conducted a human evaluation in which $100$ annotators, who were instructed to pay attention to the alignment with the given textual description and the perceived aesthetics, were asked to choose whether they preferred the initial or the optimized image based on their personal preferences. With high inter-rater reliability, the results confirm the high performance of ComfyGI with a median win rate for the image generated with the optimized workflow of approximately 90\%.

Following this introduction, Sect.~\ref{sec:background} presents related work on image optimization and GI. Section~\ref{sec:method} describes the components of ComfyGI, the used benchmarks, and the setup of the human evaluation. In Sect.~\ref{sec:experiments_results}, we present and discuss the results before concluding the paper in Sect.~\ref{sec:conclusion}. Section~\ref{sec:impact_statement} presents our broader impact statement. 

\begin{figure*}[!h]
    \centering
    \includegraphics[width=\textwidth]{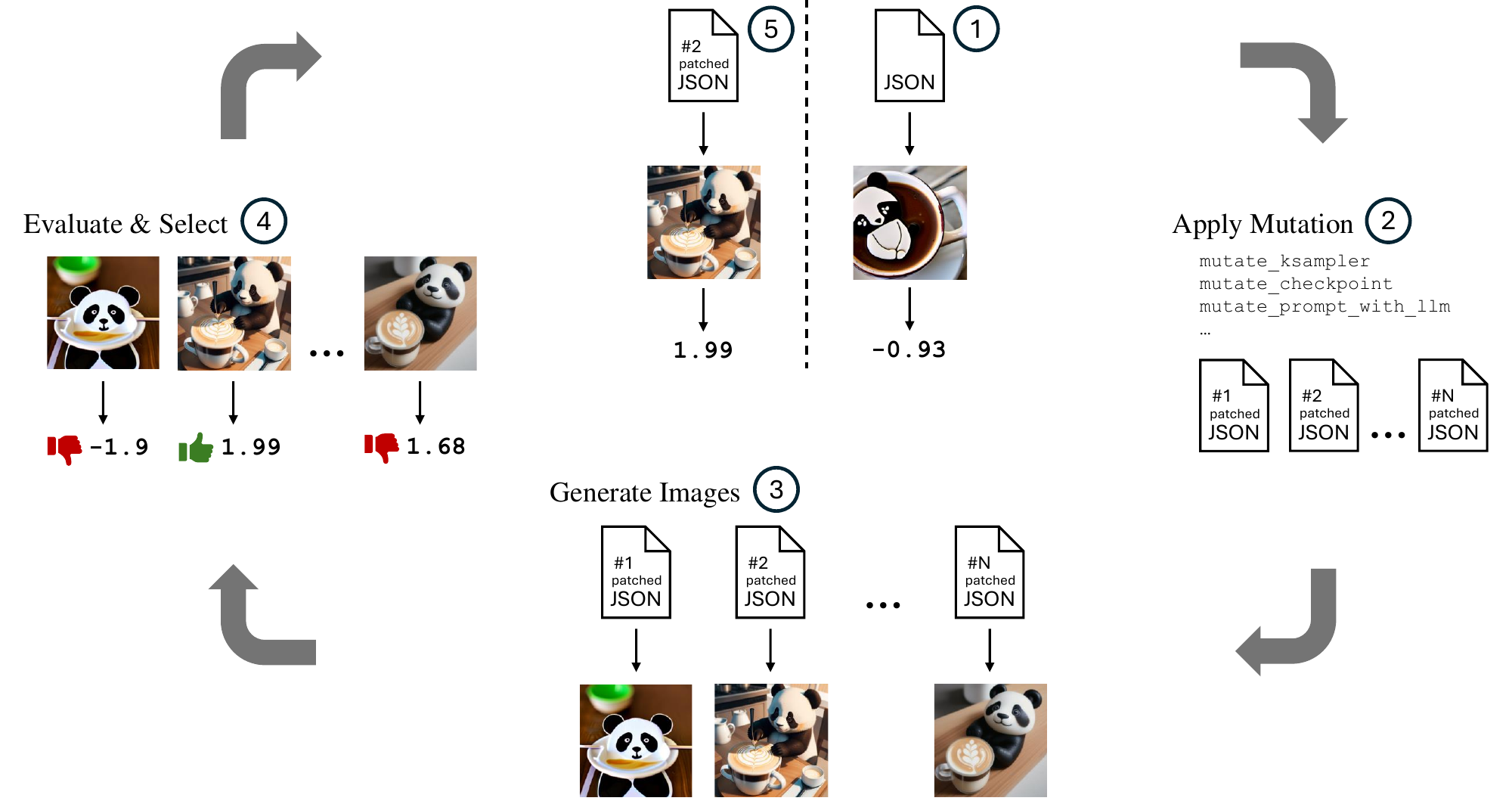}
    \caption{An illustration of ComfyGI's hill climbing method for improving workflows for text-to-image generation.}
    \label{fig:method}
\end{figure*}

\section{Background}
\label{sec:background}

In this section, we present related work on image optimization and assessment and give a brief introduction to ComfyUI as well as genetic improvement. 

\subsection{Image Optimization and Quality Assessment}

Generative text-to-image models like GANs \cite{goodfellow2020generative, reed2016generative, tao2022df}, auto-regressive models \cite{ramesh2021zero, ding2021cogview}, and diffusion models \cite{ho2020denoising,dhariwal2021diffusion,saharia2022photorealistic} can produce high quality images. Especially stable diffusion \cite{rombach2022high} has shown great performance in recent years. These models generate images based on a textual description (prompt). 
However, the quality of outputs generated by these models is highly sensitive to both the prompt as well as the hyper-parameter settings \cite{wang2022diffusiondb}. Additionally, the output might not be aligned with human preferences or not capture the intent of the user \cite{xu2024imagereward}.

Some approaches try to align the model as a whole with human preferences \cite{dong2023raft, lee2023aligning, wu2023human, xu2024imagereward} while others aim to optimize the prompt directly either with humans in the loop \cite{martins2023towards} or using automatic metrics \cite{hao2024optimizing, wang2024discrete}. Another approach uses genetic algorithms to optimize latent representations using both automatic measures as well as human interaction \cite{hall2024collaborative}.

Additionally, \citet{berger2023stableyolo} propose a method to simultaneously  optimize the prompt and hyper-parameters using a genetic algorithm. However, their work considers a quality score derived from YOLO \cite{redmon2016you} and not the alignment with users' intentions or aesthetic preferences.

Our method optimizes the entire workflow including both prompt and hyper-parameters at the same time while aligning the output with human preferences and user intent, optimizing the ImageReward score \cite{xu2024imagereward}.

\subsection{ComfyUI}

ComfyUI is a browser-based tool that can be used to create and run design workflows for automatic image generation without any programming knowledge. In these workflows, the elements required for image generation, such as prompts or sampler settings, are specified and their interaction is defined. Figure~\ref{fig:comfyui} shows how such a workflow could look like in ComfyUI. The example shows a module for defining the checkpoint model used, a module setting the image's dimensions, modules for the positive and negative prompts, as well as modules for the sampler settings and for saving the final image. The individual modules are interconnected by wires. This free combination of modules makes ComfyUI relevant also for more professional users but also easy to expand with new modules, so it is not surprising that new innovations, such as ControlNet \cite{zhang2023adding} or IP adapters \cite{ye2023ip}, are usually quickly available in ComfyUI. 

Further, the design workflows are saved in a JSON format. This allows effective workflows to be shared online with the design community, but also provides an accessible interface to automatically optimize the workflows.

\subsection{Genetic Improvement}

In software development, genetic improvement (GI) \cite{petke2017genetic} is a technique that can be used to functionally or non-functionally optimize existing software by employing search-based approaches. Functional improvement includes, e.g., automatic bug fixing \cite{yuan2020toward}, while non-functional improvement can optimize properties like runtime \cite{langdon2015improving}, memory requirements \cite{callan2022multi}, or energy consumption \cite{bruce2015reducing}. During the search, mutations, i.e. small changes to the source code, are made to change the software in order to gradually create a patch that can later be applied to the software. Mutations can, for example, insert, swap, or delete lines of code. Changes directly to the abstract syntax tree (AST) are also possible \cite{an2018comparing}, as is the integration of LLMs to generate alternative code variants within a mutation operator \cite{brownlee2023enhancing,brownlee2024large}. An objective function guides the search in the desired direction. 

GI has already been used to successfully improve software with over tens of thousands of lines of code \cite{langdon2014optimizing}. \citet{haraldsson2017fixing} present a GI approach that can also be applied in practice, in which suitable patches for a software are searched for overnight. The next day, the programmers can then choose from several suggested software patches. 

\citet{fredericks2024generativegi, fredericks2024crafting} showed that in addition to improving software, GI techniques can also be used to generate images. However, they only focused on generating images based on computational principles like flow fields, stippling, and basic geometric constructions. Photorealistic images or complex drawings and paintings, as possible with diffusion models, were not considered. 

So to our knowledge, we are the first to present an accessible approach that uses GI techniques to optimize image generation workflows and generate images that are both aligned with the given description and aesthetically appealing without human intervention.

\begin{figure*}[ht]
    \centering
    \includegraphics[width=0.797\textwidth]{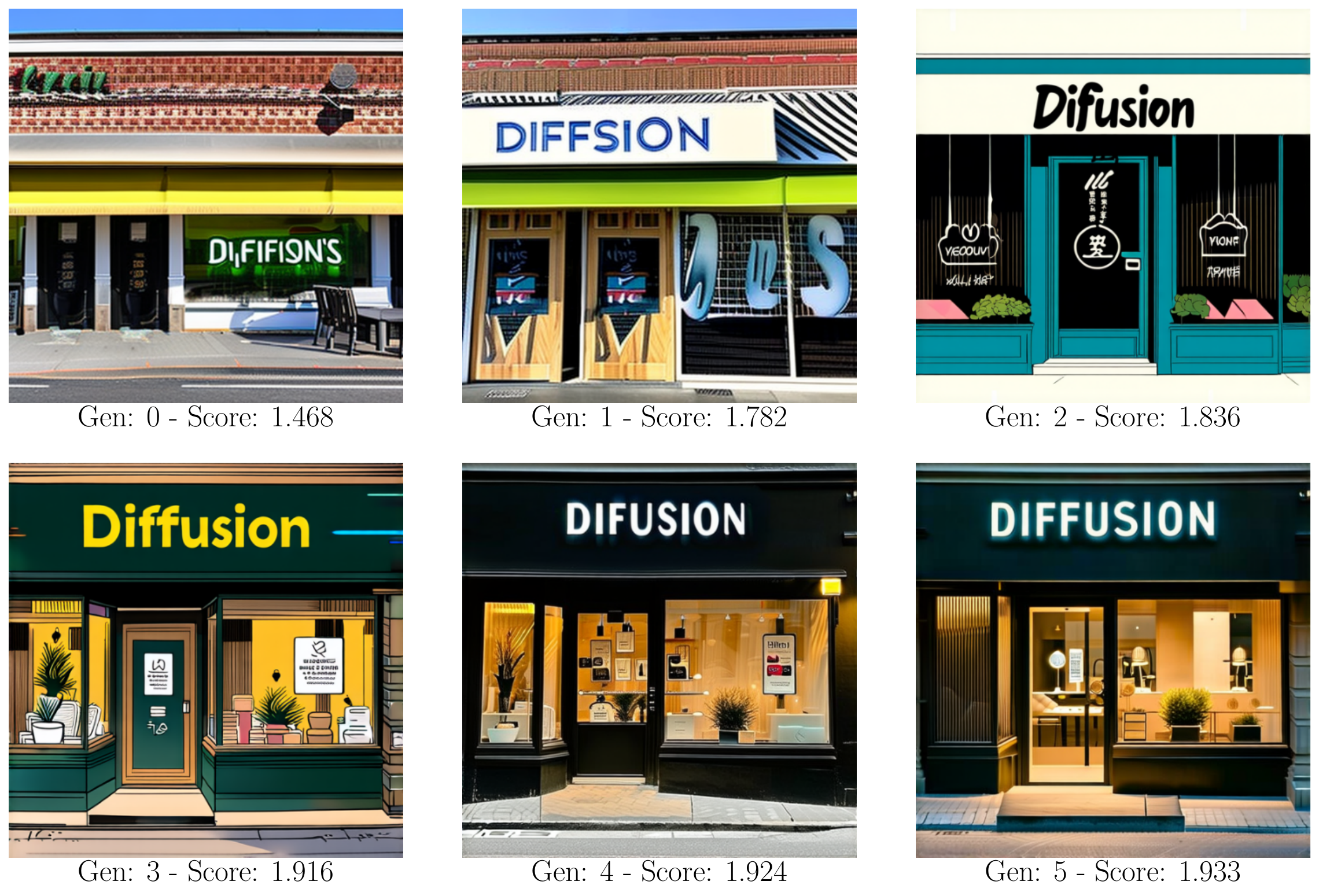}
    \caption{An example for image improvement with ComfyGI over several generations for the prompt \textit{``storefront with `diffusion' written on it''}. For every generation, we show the image and the score for the best found patch so far.}
    \label{fig:storefront_example} 
\end{figure*}

\section{Method: ComfyGI}
\label{sec:method}  

Next, we describe ComfyGI's search method including the mutation operators used, present the used diffusion models and benchmarks, and explain our approach for the human evaluation. 

\subsection{Search Method and Mutation Operators}

ComfyGI uses GI techniques to improve a given design workflow and enable the generation of a high quality image that is aligned with the input prompt and is also aesthetically appealing. To achieve this goal, we employ a hill climbing method to search for a patch that can be applied to improve the given workflow in JSON format. With this updated workflow, we generate the optimized image. 

Figure~\ref{fig:method} illustrates the steps that are iteratively performed by ComfyGI. First, we take the input workflow in JSON format, use it to generate the initial image, and assign a score to this image with the ImageReward model (step 1). In the example illustration, the assigned score is $-0.93$. After that, we search the neighborhood and apply small mutations to the workflow (step 2). These mutations range from changes to the sampling configuration to improvements of the prompts with an LLM and are explained in detail below. Following that, we generate an image from each updated workflow (step 3) and evaluate all generated images with the ImageReward model and compare the score of all images (step 4). If the score of the best image in the current generation is better than the best score recorded so far, we add the mutation that led to this successful improvement to the patch (step 5). In the example, the best image of the current generation has a score of $1.99$ and is therefore better than the best image from the previous generation (initial image with a score of $-0.93$). This process continues until no further improvement can be found. The best mutations selected up to that point per generation form the patch which is used to improve the input workflow (in JSON format). This improved workflow is then used to generate the final, optimized image. 

In our experiments, we use a text-to-image workflow with the modules and the wiring shown in Fig.~\ref{fig:comfyui}. For more details about the used workflow, including its JSON representation as well as the initial settings, we refer the reader to Appendix~\ref{app:appendix_workflow_mutation_settings}.

In order to introduce changes into such a workflow's JSON representation, we designed a variety of mutation operators specifically for the optimization of image generation workflows. Each mutation operator applies the changes to a specific module of the workflow. The mutation operators used in the experiments are the following:
\begin{itemize}
    \item \texttt{\textbf{checkpoint}}: The workflow's checkpoint model can be randomly replaced by another one chosen from a pre-defined set. 
    \item \texttt{\textbf{ksampler}}: Changes the relevant settings of the \textit{ksampler} module from ComfyUI. Whenever this mutation operator is called, a single property is randomly changed. Supported are the seed, the number of steps, the classifier free guidance (CFG), the sampler, the scheduler, as well as the denoise level. For numerical values we specified a reasonable range of values and for the sampler and scheduler we provided a list of possible values.
    \item \texttt{\textbf{prompt\_word}}: Changes the text of the prompt modules (positive and negative prompt) by either randomly \textit{removing}, \textit{switching}, or \textit{copying} a word from the existing prompt. The procedure is similar to classic operators from GI in software development, where lines of code can, e.g., be moved or deleted. 
    \item \texttt{\textbf{prompt\_statement}}: Works like \texttt{prompt\_word} but focuses on what we call prompt statements -- larger parts of the prompt that are separated by a comma. In addition to \texttt{prompt\_word}'s operators, also \textit{add} and \textit{replace} operations are supported, which allow to integrate phrases from a pre-defined list into the prompt. This enables including common expressions such as ``digital painting'' or ``ultra realistic''. The operator distinguishes between positive and negative prompts.
    \item \texttt{\textbf{prompt\_llm}}: Requests an LLM to optimize the current prompt. The used LLM model as well as its seed and temperature are randomly determined by the \texttt{prompt\_llm} mutation operator. Further, we provide different prompt templates for the request, depending on whether a positive or negative prompt should be optimized.
\end{itemize}
As mentioned above, the models used by the \texttt{checkpoint} mutation operator can be easily defined. The models we use in our experiments are mentioned in Sect.~\ref{sec:models_benchmarks}. The \texttt{prompt\_statement} operator can enrich the workflow's prompts with additional statements or replace existing ones. To make this possible, in our experiments we provide the mutation operator with the $250$ most common prompt statements that we extracted from a large prompt collection \cite{santana2022} for the positive prompts. For the negative prompts, we took statements from \citet{yip2023}. For the \texttt{prompt\_llm} operator, we use LLMs provided by Ollama\footnote{\url{https://ollama.com/}}, namely: \textit{llama3.1:8b}, \textit{mistral-nemo:12b}, and \textit{gemma2:9b}.

Further implementation details of the mutation operators and the method in general are presented in Appendix~\ref{app:appendix_workflow_mutation_settings}.

\begin{figure}[ht!]
    \centering
    \subfloat[Initial prompt: \textit{``a panda making latte art''}]{
        \includegraphics[width=\columnwidth]{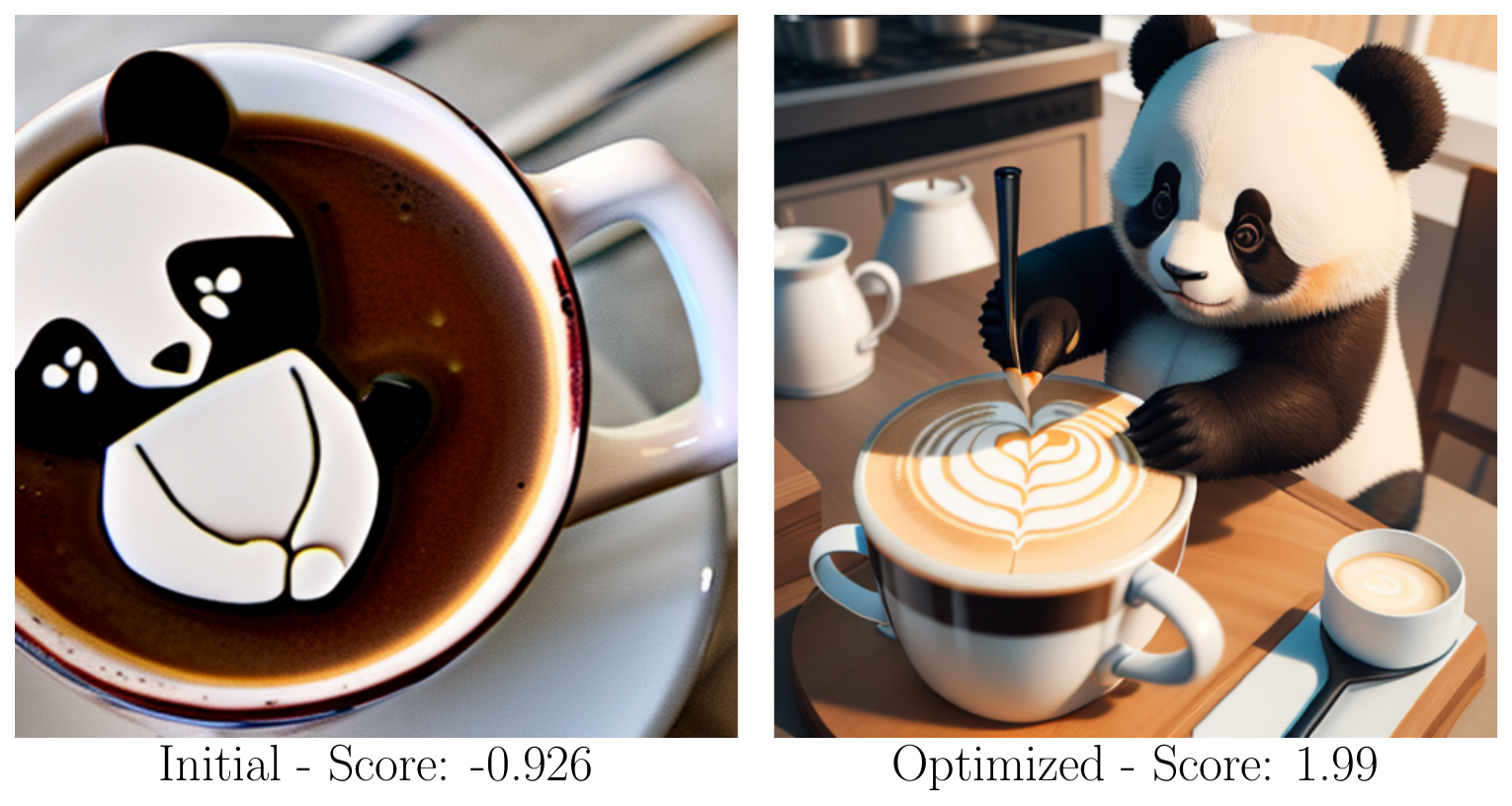}
        \label{fig:panda_latte_art}
    }\hfill
    \subfloat[Initial prompt: \textit{``mcdonalds church''}]{
        \includegraphics[width=\columnwidth]{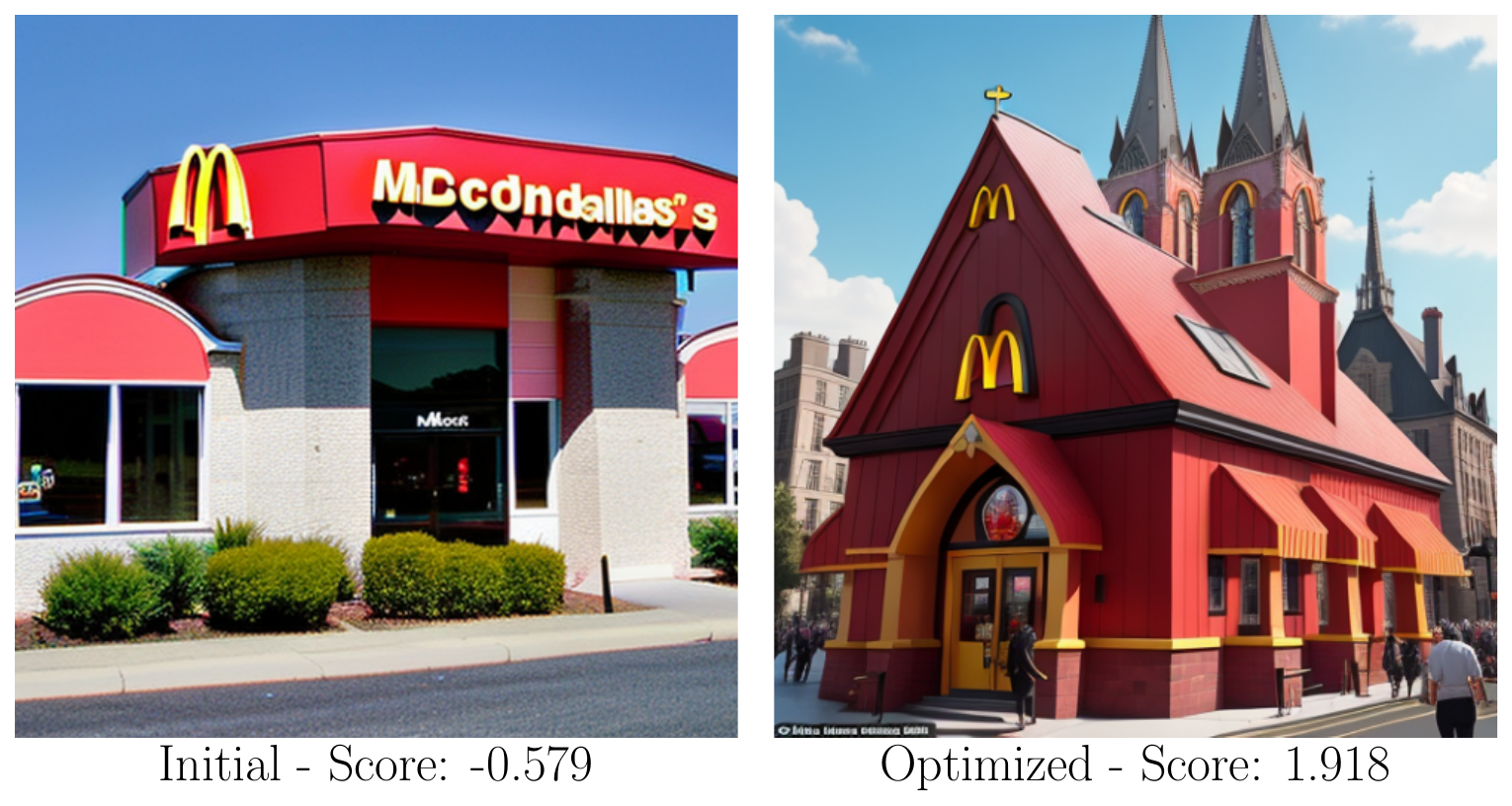}
        \label{fig:mcdonalds_church}
    }\hfill
    \subfloat[Initial prompt: \textit{``two cars on the street''}]{
        \includegraphics[width=\columnwidth]{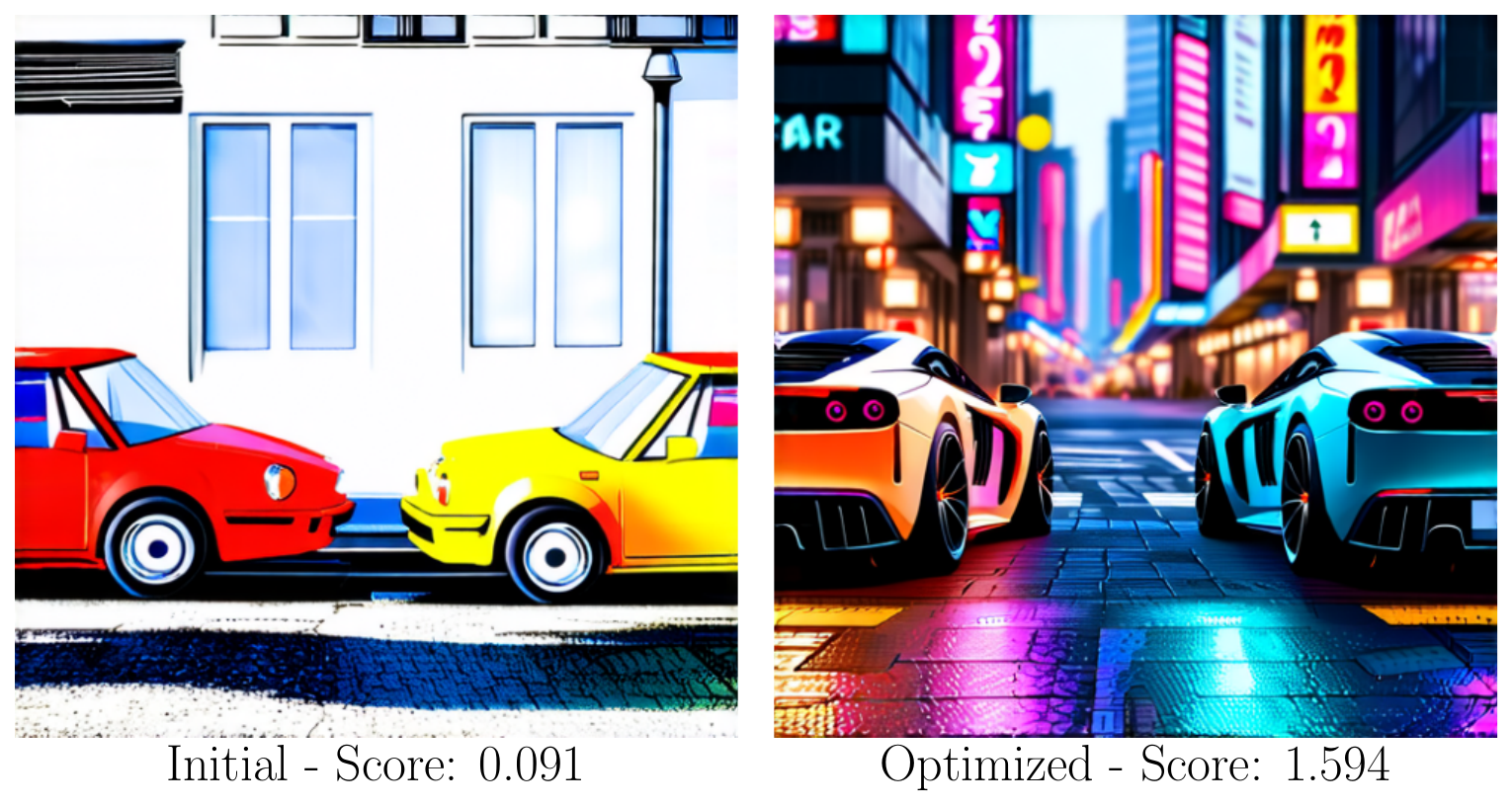}
        \label{fig:two_cars}
    }
    \caption{Three examples for image improvement with ComfyGI. The left image shows the initial image and the right one the optimized counterpart.}
    \label{fig:run_examples}
\end{figure}

\subsection{Diffusion Models and Benchmarks}
\label{sec:models_benchmarks}

The trained diffusion models are the foundation for every design workflow. In our experiments, we use models that are freely available on Hugging Face. When making the selection, we paid attention to select a broad variety of models and took popularity and the number of previous downloads into account. The models used in the experiments are: \textit{Stable Diffusion 1.5}, \textit{Stable Diffusion 2}, \textit{Stable Diffusion 3 Medium}, \textit{Stable Diffusion XL Turbo 1.0}, \textit{Stable Diffusion XL Base 1.0}, \textit{Realistic Vision 6.0}, \textit{ReV Animated 1.2.2}, \textit{Dreamlike Photoreal 2.0}, and \textit{DreamShaper 3.3}. Further details on the models are given in Appendix~\ref{app:model_versions}. 

To evaluate ComfyGI's performance, we randomly sampled $42$ benchmark prompts from all $14$ categories ($3$ prompts per category) suggested for text-to-image generation by ImagenHub \cite{kuimagenhub} for our experiments. The prompt categories range, for example, from rare words and misspellings to text generation in images. An overview of the categories and associated prompts is given in Appendix~\ref{app:benchmark_problems}.

\subsection{Human Evaluation}

We use the ImageReward model \cite{xu2024imagereward} to guide ComfyGI's search process, which also gives us an evaluation for the generated images, as the model was trained to assess images based on their aesthetics and their alignment with the prompt. However, the question arises whether potential users see it the same way? For verification testing, we also carried out a human evaluation of our results. 

We asked participants to assess given image pairs -- the initially generated image vs. the optimized image -- and decide which of them they find to align better with the given description and is more aesthetically appealing from their perspective. To prepare the participants for this task, we carried out a priming in which we explained them that they should first pay attention to the alignment with the description. If only one of the images fits the description, then that one should be selected. If both fit, the perceived aesthetics of the images should also be taken into account. If neither of the two images fit the description, the perceived aesthetics of the images should be decisive for the selection. Further, three additional image pairs were displayed to the participants during the study as attention checks. Those three image pairs were clearly distinguishable and only one image (per pair) was aligned with the description. Participants who did not pass one or more of these checks were excluded from the analysis of the results.

After the assessment of the image pairs, we collected demographic information from the participants and also asked for their proficiency of the English language and their knowledge on image generation.

For more details on the priming, the attention checks, and the design of the questionnaire, we refer the reader to Appendix~\ref{app:human_eval}.

For the recruitment of participants, we chose the Prolific\footnote{\url{https://www.prolific.com/}} platform for our human evaluation. Crucial factors for this decision were the expected data quality \cite{peer2017beyond} as well as the platform's high ethical standards. 

\section{Experiments and Results}
\label{sec:experiments_results}

\begin{figure*}[ht]
    \centering
    \begin{minipage}[b]{0.226\textwidth}
        \centering
        \includegraphics[width=\textwidth]{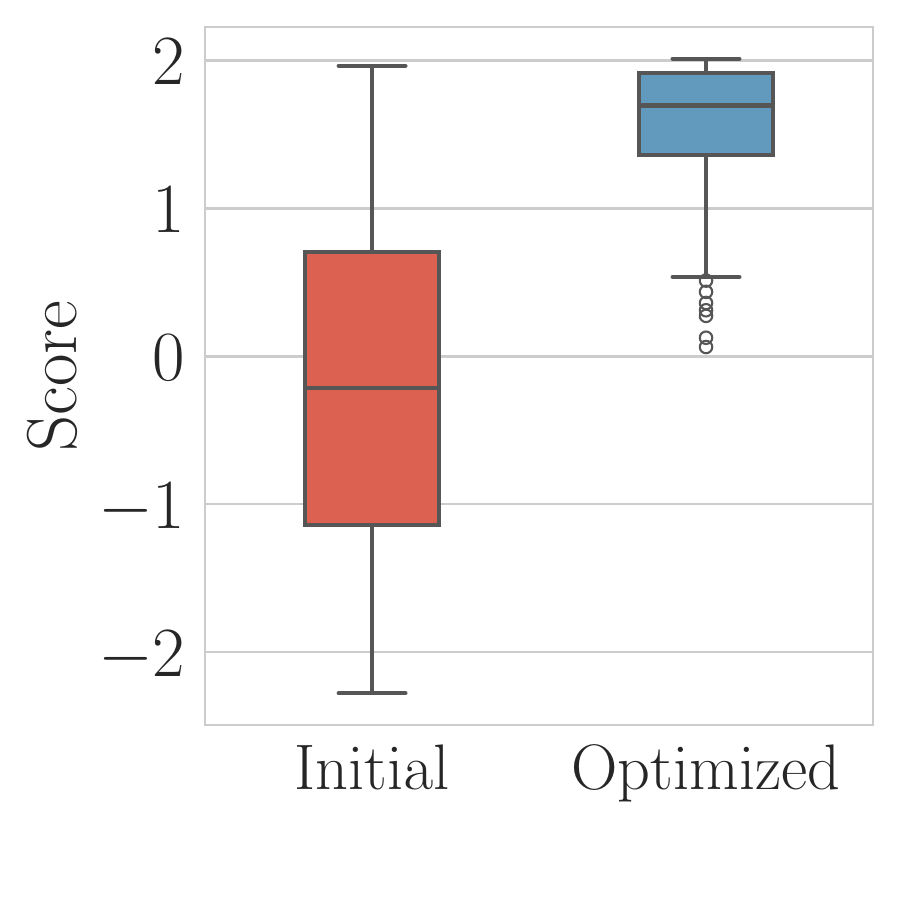}
        \caption{Scores for the initial and optimized images.}
        \label{fig:scores_box_plot1}
    \end{minipage}
    \hfill
    \begin{minipage}[b]{0.375\textwidth}
        \centering
        \includegraphics[width=\textwidth]{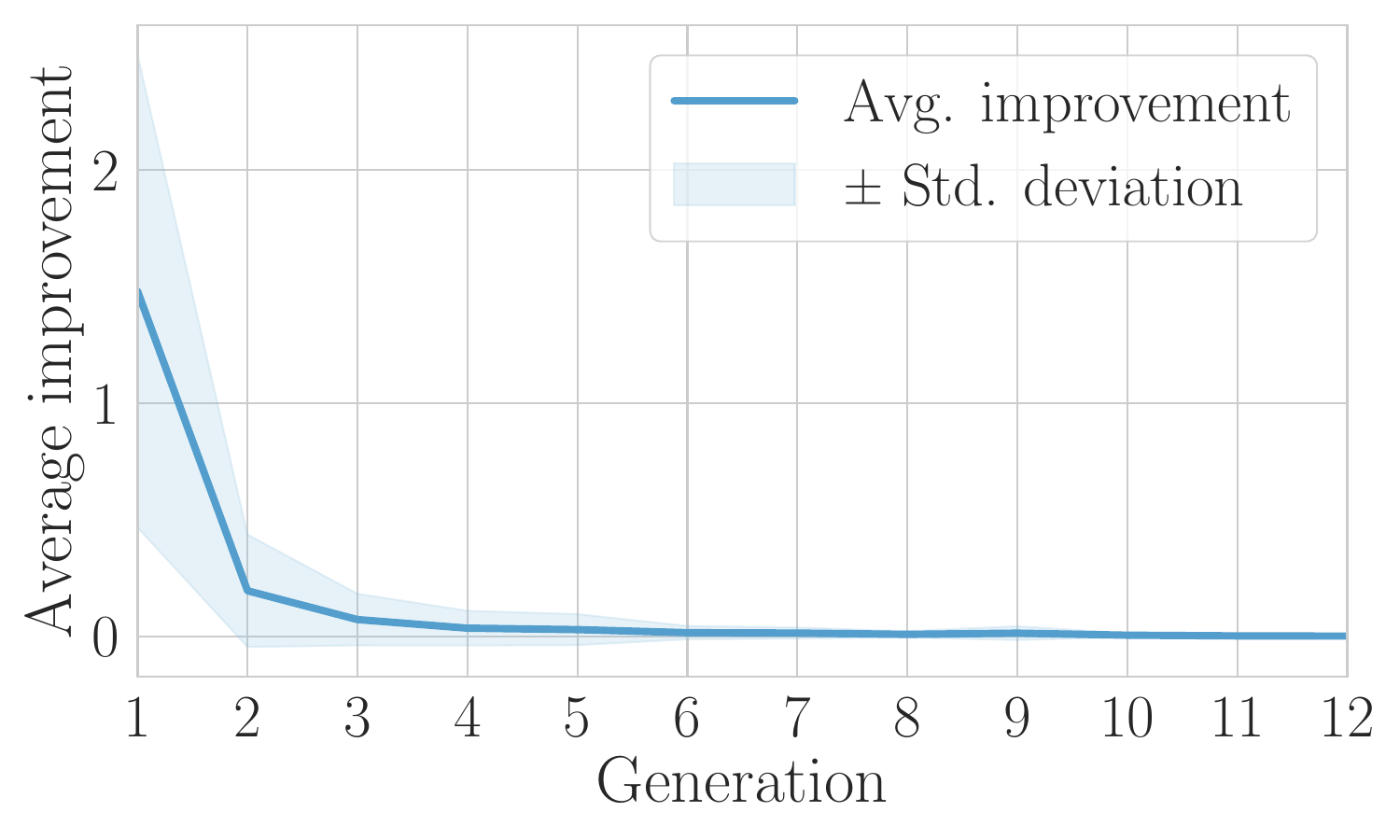}
        \caption{Average improvement and standard deviation of the ImageReward score over generations.}
        \label{fig:improvement_line_plot}
    \end{minipage}
    \hfill
    \begin{minipage}[b]{0.375\textwidth}
        \centering
        \includegraphics[width=\textwidth]{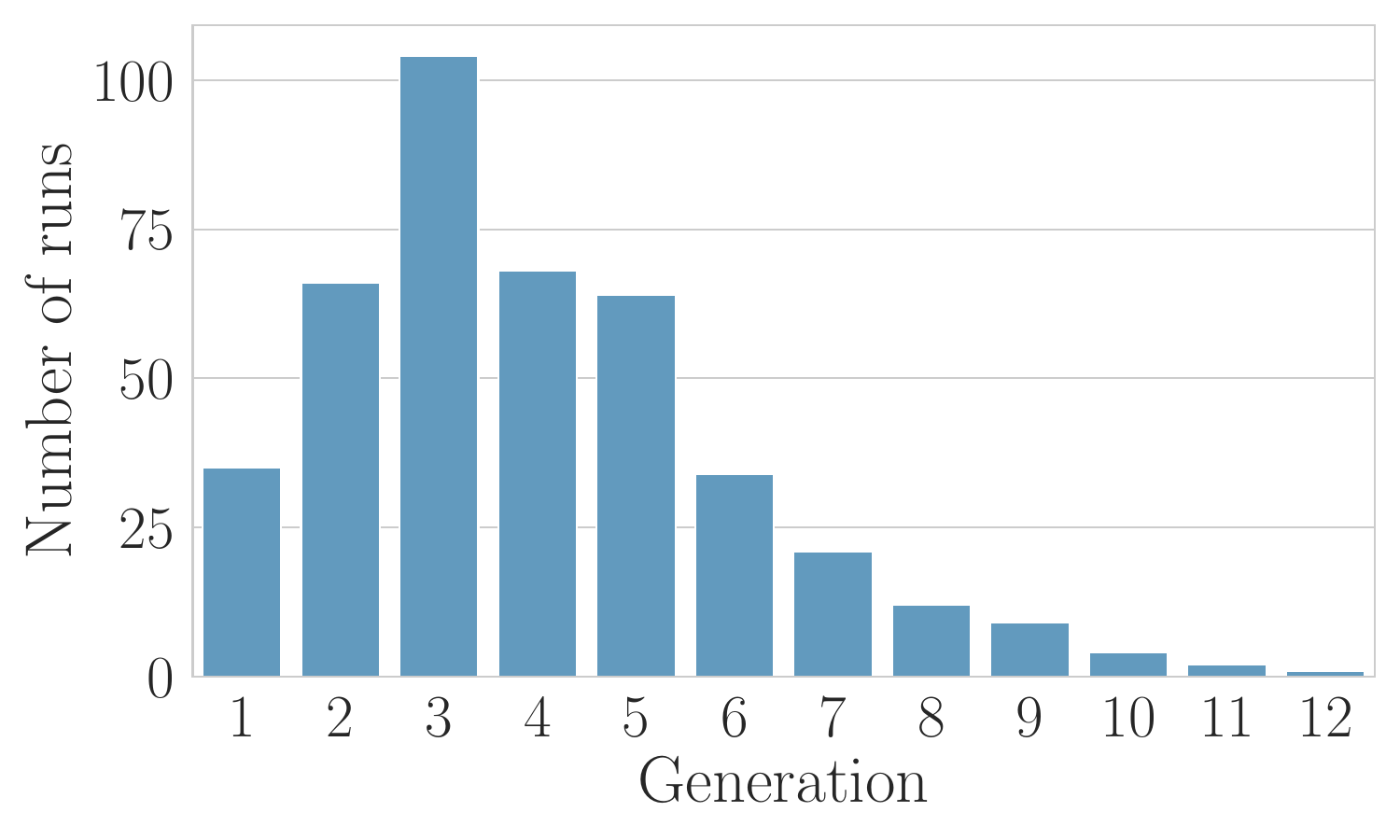}
        \caption{Number of generations till convergence for all runs.}
        \label{fig:generation_numbers}
    \end{minipage}
\end{figure*}

In this section, we present and discuss the results achieved with ComfyGI, including visual examples as well as the outcome of the human evaluation.

\subsection{Results of the ComfyGI Runs}
\label{sec:results_with_score}

To assess the performance of ComfyGI, we generated and evaluated images for $42$ prompts out of $3$ categories from the ImagenHub \cite{kuimagenhub} benchmark. To obtain reliable results despite the nondeterministic nature of the image generation models, we performed $10$ independent runs for every benchmark prompt. For each of the $10$ runs we always used a random checkpoint model and a random seed in the initial workflow. In addition, in each generation we consider $30$ neighboring solutions for each mutation operator used for a total of $150$ possible different mutations per generation.

First, we start with a visual inspection of some example images optimized with ComfyGI. Figure~\ref{fig:storefront_example} shows the improvement over several generations for the initial prompt \textit{``storefront with `diffusion' written on it''}. For every generation, the image and the score for the best found patch so far is shown. We see that the initial image (Gen. $0$ with a score of $1.468$) contains many errors and the text on the storefront is hard to read and misspelled. Over the generations, the images get better and better. The structures become clearer and the colors get brighter and more expressive. In the final image (Gen. 5) with a score of $1.933$, even the text on the storefront is written correctly. This was achieved by applying, among others, the mutation operators \texttt{prompt\_llm}, \texttt{checkpoint} and \texttt{ksampler}.

Figure~\ref{fig:run_examples} shows further examples of generated images. The left-hand images show the initially generated image and the right-hand images show the counterpart optimized with ComfyGI. The intermediary steps are omitted due to space limitations but are shown in Appendix~\ref{app:images_intermediary} for the interested reader. As before, we see that the optimized images (images on the right), are aesthetically more appealing, the structures are clearer, the colors more vibrant, which is also supported by a significantly better score. For example, the score of the optimized image in Fig.~\ref{fig:two_cars} is more than $1.5$ points higher than that of the initial image. The optimized image (right) also shows a modern, colorful street in the background while at the same time playing with beautiful light reflections, while the initial image (left) shows a very flat-looking illustration. In addition to the perceived aesthetics, we see especially in Figs.~\ref{fig:panda_latte_art} (initial prompt: \textit{``a panda making latte art''}) and \ref{fig:mcdonalds_church} (initial prompt: \textit{``mcdonalds church''}) that also the alignment with the given prompt is optimized. In the optimized image in Fig.~\ref{fig:panda_latte_art} we can see a panda in the process of making latte art. The optimized image in Fig.~\ref{fig:mcdonalds_church} shows a church with the well-known logo over the entrance. 

Second, we analyze the overall performance using the ImageReward score. Figure~\ref{fig:scores_box_plot1} shows box-plots for the achieved scores of the initial and the optimized images for all $420$ image pairs ($10$ runs for $42$ benchmark prompts). We see that the median ImageReward score could be significantly improved by about 50\% compared to the initial images. In addition, also the variance of the score is lower for the optimized images. These findings also hold for each studied prompt category (see Fig.~\ref{fig:score_box_plots_by_category} in Appendix~\ref{app:additional_analyses_image_generation}).

To study how this improvement is distributed over the generations of the search or in other words, how long it takes for the search to converge, we analyze the average improvement per generation. Figure~\ref{fig:improvement_line_plot} shows the average improvement and the standard deviation of the ImageReward score over the generations of the search. We see the largest improvements in the first $3$ generations. Afterwards, only slight improvements are found. Connected to that, shows Fig.~\ref{fig:generation_numbers} the number of generations that were used by the hill climber. We see that most of the runs took $3$ generations to converge and only a minority of runs took $6$ or more generations to converge.
So the success we can see in Fig.~\ref{fig:scores_box_plot1} can be achieved already in a low number of generations.

\subsection{Results of the Human Evaluation}

We also carried out a human evaluation study to further confirm our findings. Therefore, we recruited $100$ participants on the Prolific platform -- $10$ participants for each performed run. Every participant had to evaluate $42$ image pairs.\footnote{Four out of the total $420$ image pairs were not presented to the participants because they contained content that is not suitable for all audiences. However, these issues only existed for the initially generated images. The images optimized by ComfyGI did not contain any issues of this nature.} Participants took 12:41 minutes median time to complete the study. This resulted in a median hourly wage of \pounds$14.19$. 

The participants were 57\% male and 43\% female with a median age of 39 years. In addition, all participants rated their knowledge of the English language as very good. Overall, one participant failed the attention checks and was therefore not included in the results. For further details on the demographics, we refer the reader to Appendix~\ref{app:results_human_evaluation}.

\begin{figure}[h!]
    \centering
    \includegraphics[width=0.9\linewidth]{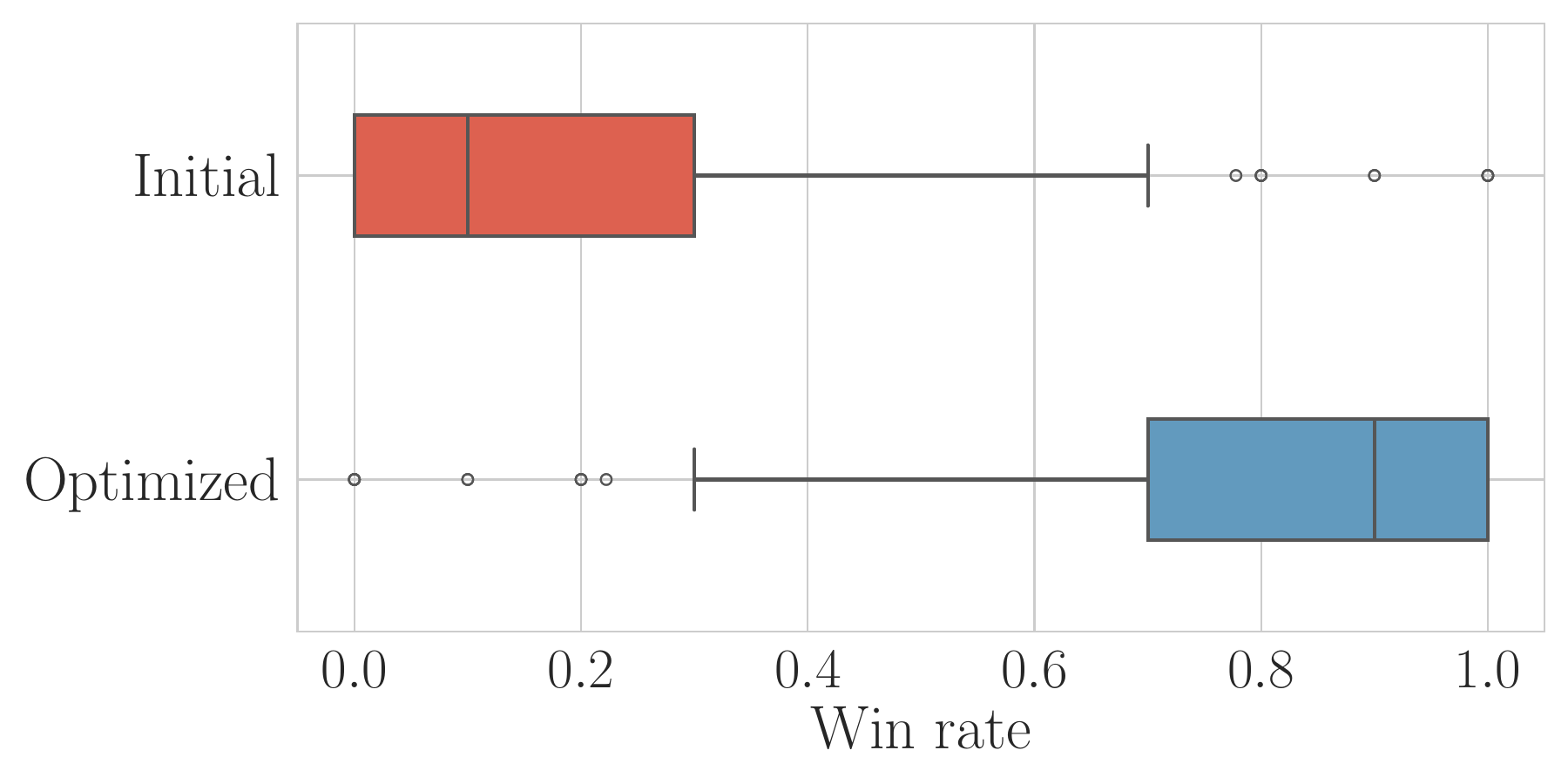}
    \caption{Win rate of the initial and optimized images for all prompts and runs in the human evaluation.}
    \label{fig:humaneval_boxplot_overall}
\end{figure}

\begin{figure*}[!h]
    \centering
    \begin{minipage}[b]{0.49\textwidth}
        \centering
        \includegraphics[width=\textwidth]{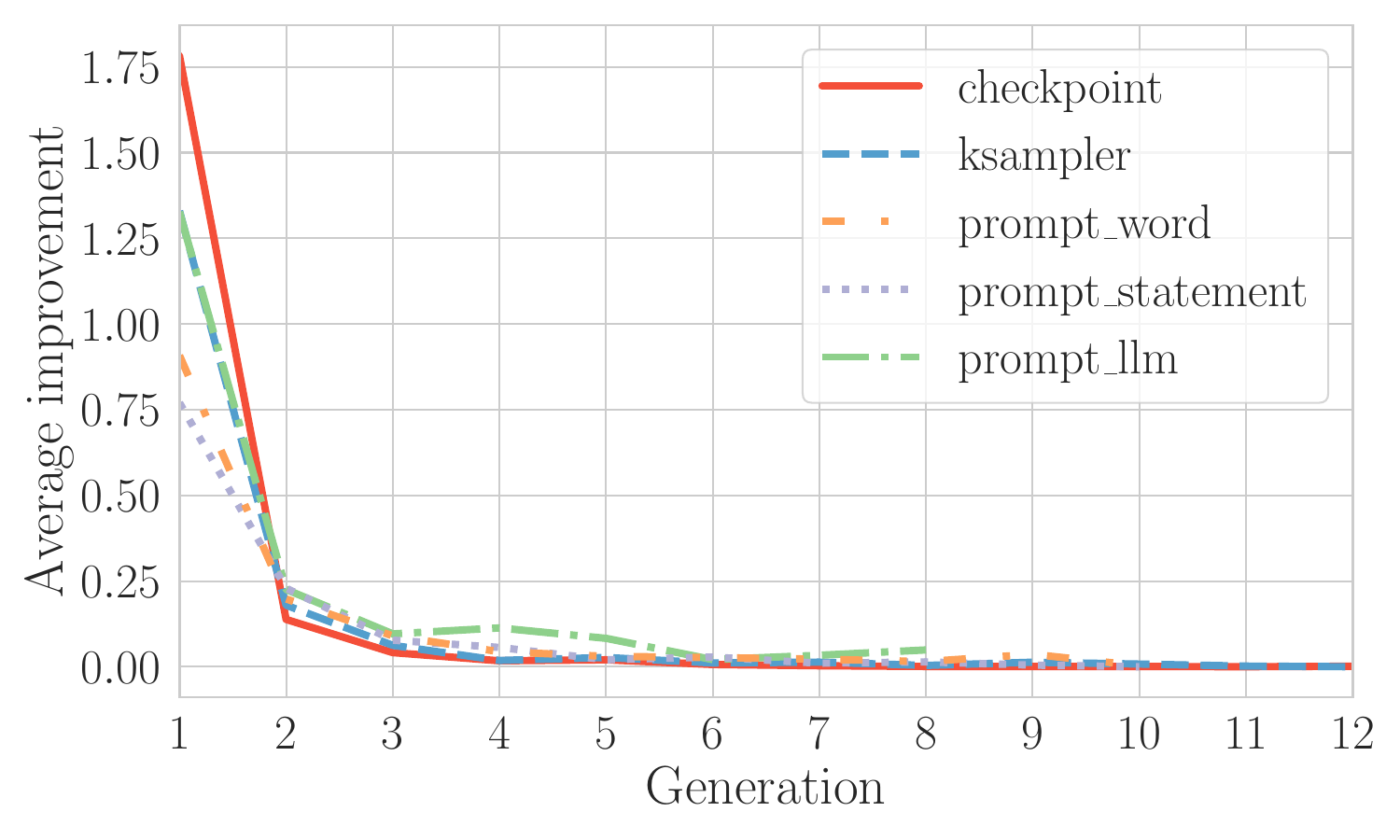}
        \caption{Average improvement of the ImageReward score over generations for all mutation variants.}
        \label{fig:patch_improvement_line_plot}
    \end{minipage}
    \hfill
    \begin{minipage}[b]{0.49\textwidth}
        \centering
        \includegraphics[width=\textwidth]{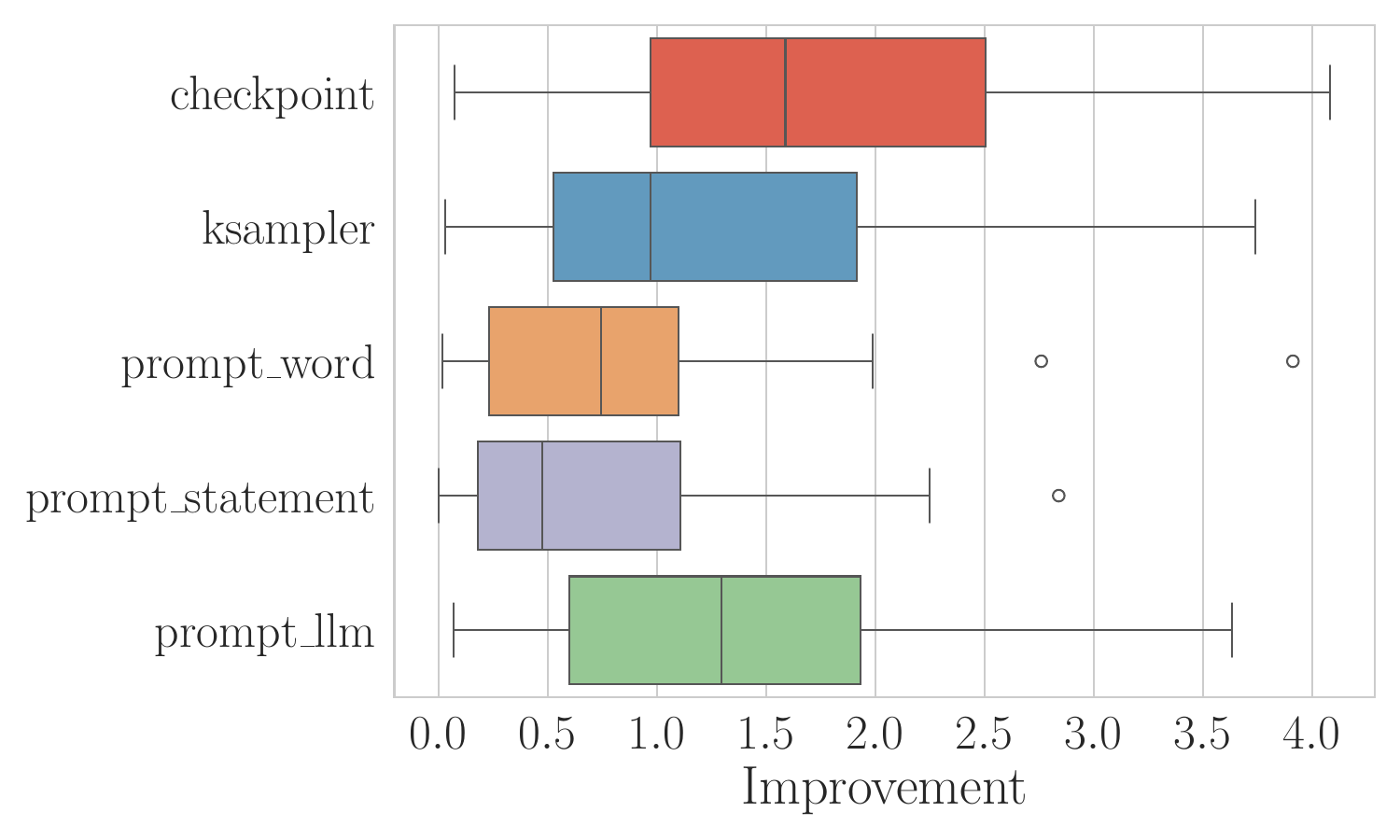}
        \caption{Box-plots for the applied mutations in the first generation over all prompts and runs.}
        \label{fig:patch_improvement_box_plot_first_gen}
    \end{minipage}
\end{figure*}

Figure~\ref{fig:humaneval_boxplot_overall} shows box-plots for the win rate in the human evaluation for the initial as well as for the optimized images, where the win rate is the proportion of human annotators that prefer an image (initial or optimized) in a direct comparison to its counterpart. We see that the human evaluation confirms the previous findings as the optimized images were preferred in about 90\% of the cases. This difference is also significant according to a Wilcoxon signed-rank test \cite{wilcoxon1992individual} with a \textit{p}-value $<0.0001$. As mentioned before for the ImageReward score, also the positive results of the human evaluation hold for all studied prompt categories (see Fig.~\ref{fig:humaneval_box_plots_by_category} in Appendix~\ref{app:results_human_evaluation}). We also checked for inter-rater reliability (IRR) using Gwet’s AC1 \cite{gwet2008computing,gwet2014handbook} to counteract the ``paradox of kappa'' in high agreement scenarios \cite{cicchetti1990high}. For Gwet's AC1 coefficient we calculated a value of $0.63459$ ensuring high IRR.

\subsection{Influence of the Mutation Operators}

In addition to the objective function, the mutation operators are the driving factors of ComfyGI's guided search. Consequently, we study their influence on the overall improvement. 

Figure~\ref{fig:patch_improvement_line_plot} shows the average improvement of the ImageReward score over all studied prompts and runs. As before in Sect.~\ref{sec:results_with_score}, we see that the improvement is most effective in the first generations. For example, we see an average improvement of over $1.75$ points with the \texttt{checkpoint} mutation operator in the first generation. In addition also the \texttt{ksampler} and \texttt{prompt\_llm} operators perform very well in the first generation. Especially the \texttt{prompt\_llm} operator is also important in the following generations (see generations 3-5). 

As the biggest improvements are found in the first generation, we take a closer look at these improvements. Figure~\ref{fig:patch_improvement_box_plot_first_gen} shows box-plots for the applied mutations in the first generation over all runs and considered benchmark prompts. As before, we see that the \texttt{checkpoint} operator has the highest impact. So changing the checkpoint model in the first generations plays a crucial role. However, it is by no means the case that the most advanced model is always selected, which in our case would be \textit{Stable Diffusion 3 Medium}. This depends heavily on the image to be generated. For example, for the categories \textit{Rare Words} and \textit{Text}, the \textit{Stable Diffusion 3 Medium} model is actually used most frequently. But, e.g., for the categories \textit{Conflicting} and \textit{Counting}, the checkpoint models \textit{Realistic Vision 6.0}, and \textit{ReV Animated 1.2.2} are used more frequently (for more details see Fig.~\ref{fig:model_bar_plot_by_category} in Appendix~\ref{app:additional_analyses_image_generation}). As above, also the \texttt{prompt\_llm}~operator is very important, as it allows to adjust the prompts for the workflow at hand which enables more detailed prompts than given by the ImagenHub benchmark set.

\section{Conclusion}
\label{sec:conclusion}

In this paper, we introduced ComfyGI, an approach that uses GI techniques to automatically improve workflows for image generation without the need for human intervention. This allows images to be generated with a significantly higher quality in terms of the output image's alignment with the given description and its perceived aesthetics. 

In our analysis of ComfyGI's performance, we found that overall, the images generated with an optimized workflow are about 50\% better than with the initial workflow in terms of the median ImageReward score. This was also confirmed by a human evaluation with $100$ participants as the improved images where preferred in about 90\% of the cases. 

In future work, we will investigate more complex workflows and additional mutation operators, due to the easy extensibility of ComfyGI.

\section{Impact Statement}
\label{sec:impact_statement}

ComfyGI makes it easier to use image design workflows, which could have a positive social impact through increased inclusivity. In its current form, we therefore do not see any negative impact. However, due to the simple extensibility of ComfyGI, the objective function could, e.g., be changed and used for other potentially negative purposes. We therefore call for a careful and respectful usage.

\bibliography{example_paper}

\begin{thebibliography}{44}
\providecommand{\natexlab}[1]{#1}
\providecommand{\url}[1]{\texttt{#1}}
\expandafter\ifx\csname urlstyle\endcsname\relax
  \providecommand{\doi}[1]{doi: #1}\else
  \providecommand{\doi}{doi: \begingroup \urlstyle{rm}\Url}\fi

\bibitem[An et~al.(2018)An, Kim, and Yoo]{an2018comparing}
An, G., Kim, J., and Yoo, S.
\newblock Comparing line and ast granularity level for program repair using pyggi.
\newblock In \emph{Proceedings of the 4th International Workshop on Genetic Improvement Workshop}, pp.\  19--26, 2018.

\bibitem[Berger et~al.(2023)Berger, Dakhama, Ding, Even-Mendoza, Kelly, Menendez, Moussa, and Sarro]{berger2023stableyolo}
Berger, H., Dakhama, A., Ding, Z., Even-Mendoza, K., Kelly, D., Menendez, H., Moussa, R., and Sarro, F.
\newblock Stableyolo: Optimizing image generation for large language models.
\newblock In \emph{International Symposium on Search Based Software Engineering}, pp.\  133--139. Springer, 2023.

\bibitem[Brownlee et~al.(2023)Brownlee, Callan, Even-Mendoza, Geiger, Hanna, Petke, Sarro, and Sobania]{brownlee2023enhancing}
Brownlee, A.~E., Callan, J., Even-Mendoza, K., Geiger, A., Hanna, C., Petke, J., Sarro, F., and Sobania, D.
\newblock Enhancing genetic improvement mutations using large language models.
\newblock In \emph{International Symposium on Search Based Software Engineering}, pp.\  153--159. Springer, 2023.

\bibitem[Brownlee et~al.(2024)Brownlee, Callan, Even-Mendoza, Geiger, Hanna, Petke, Sarro, and Sobania]{brownlee2024large}
Brownlee, A. E.~I., Callan, J., Even-Mendoza, K., Geiger, A., Hanna, C., Petke, J., Sarro, F., and Sobania, D.
\newblock Large language model based mutations in genetic improvement.
\newblock 2024.

\bibitem[Bruce et~al.(2015)Bruce, Petke, and Harman]{bruce2015reducing}
Bruce, B.~R., Petke, J., and Harman, M.
\newblock Reducing energy consumption using genetic improvement.
\newblock In \emph{Proceedings of the 2015 Annual Conference on Genetic and Evolutionary Computation}, pp.\  1327--1334, 2015.

\bibitem[Callan \& Petke(2022)Callan and Petke]{callan2022multi}
Callan, J. and Petke, J.
\newblock Multi-objective genetic improvement: A case study with evosuite.
\newblock In \emph{International Symposium on Search Based Software Engineering}, pp.\  111--117. Springer, 2022.

\bibitem[Cicchetti \& Feinstein(1990)Cicchetti and Feinstein]{cicchetti1990high}
Cicchetti, D.~V. and Feinstein, A.~R.
\newblock High agreement but low kappa: Ii. resolving the paradoxes.
\newblock \emph{Journal of clinical epidemiology}, 43\penalty0 (6):\penalty0 551--558, 1990.

\bibitem[Dhariwal \& Nichol(2021)Dhariwal and Nichol]{dhariwal2021diffusion}
Dhariwal, P. and Nichol, A.
\newblock Diffusion models beat gans on image synthesis.
\newblock \emph{Advances in neural information processing systems}, 34:\penalty0 8780--8794, 2021.

\bibitem[Ding et~al.(2021)Ding, Yang, Hong, Zheng, Zhou, Yin, Lin, Zou, Shao, Yang, et~al.]{ding2021cogview}
Ding, M., Yang, Z., Hong, W., Zheng, W., Zhou, C., Yin, D., Lin, J., Zou, X., Shao, Z., Yang, H., et~al.
\newblock Cogview: Mastering text-to-image generation via transformers.
\newblock \emph{Advances in neural information processing systems}, 34:\penalty0 19822--19835, 2021.

\bibitem[Dong et~al.(2023)Dong, Xiong, Goyal, Zhang, Chow, Pan, Diao, Zhang, Shum, and Zhang]{dong2023raft}
Dong, H., Xiong, W., Goyal, D., Zhang, Y., Chow, W., Pan, R., Diao, S., Zhang, J., Shum, K., and Zhang, T.
\newblock Raft: Reward ranked finetuning for generative foundation model alignment.
\newblock \emph{arXiv preprint arXiv:2304.06767}, 2023.

\bibitem[Fredericks et~al.(2024{\natexlab{a}})Fredericks, Bobeldyk, and Moore]{fredericks2024crafting}
Fredericks, E.~M., Bobeldyk, D., and Moore, J.~M.
\newblock Crafting generative art through genetic improvement: Managing creative outputs in diverse fitness landscapes.
\newblock \emph{arXiv preprint arXiv:2407.20095}, 2024{\natexlab{a}}.

\bibitem[Fredericks et~al.(2024{\natexlab{b}})Fredericks, Moore, and Diller]{fredericks2024generativegi}
Fredericks, E.~M., Moore, J.~M., and Diller, A.~C.
\newblock Generativegi: creating generative art with genetic improvement.
\newblock \emph{Automated Software Engineering}, 31\penalty0 (1):\penalty0 23, 2024{\natexlab{b}}.

\bibitem[Goodfellow et~al.(2020)Goodfellow, Pouget-Abadie, Mirza, Xu, Warde-Farley, Ozair, Courville, and Bengio]{goodfellow2020generative}
Goodfellow, I., Pouget-Abadie, J., Mirza, M., Xu, B., Warde-Farley, D., Ozair, S., Courville, A., and Bengio, Y.
\newblock Generative adversarial networks.
\newblock \emph{Communications of the ACM}, 63\penalty0 (11):\penalty0 139--144, 2020.

\bibitem[Gwet(2008)]{gwet2008computing}
Gwet, K.~L.
\newblock Computing inter-rater reliability and its variance in the presence of high agreement.
\newblock \emph{British Journal of Mathematical and Statistical Psychology}, 61\penalty0 (1):\penalty0 29--48, 2008.

\bibitem[Gwet(2014)]{gwet2014handbook}
Gwet, K.~L.
\newblock \emph{Handbook of inter-rater reliability: The definitive guide to measuring the extent of agreement among raters}.
\newblock Advanced Analytics, LLC, 2014.

\bibitem[Hall \& Yaman(2024)Hall and Yaman]{hall2024collaborative}
Hall, O. and Yaman, A.
\newblock Collaborative interactive evolution of art in the latent space of deep generative models.
\newblock In \emph{International Conference on Computational Intelligence in Music, Sound, Art and Design (Part of EvoStar)}, pp.\  194--210. Springer, 2024.

\bibitem[Hao et~al.(2024)Hao, Chi, Dong, and Wei]{hao2024optimizing}
Hao, Y., Chi, Z., Dong, L., and Wei, F.
\newblock Optimizing prompts for text-to-image generation.
\newblock \emph{Advances in Neural Information Processing Systems}, 36, 2024.

\bibitem[Haraldsson et~al.(2017)Haraldsson, Woodward, Brownlee, and Siggeirsdottir]{haraldsson2017fixing}
Haraldsson, S.~O., Woodward, J.~R., Brownlee, A.~E., and Siggeirsdottir, K.
\newblock Fixing bugs in your sleep: How genetic improvement became an overnight success.
\newblock In \emph{Proceedings of the Genetic and Evolutionary Computation Conference Companion}, pp.\  1513--1520, 2017.

\bibitem[Ho et~al.(2020)Ho, Jain, and Abbeel]{ho2020denoising}
Ho, J., Jain, A., and Abbeel, P.
\newblock Denoising diffusion probabilistic models.
\newblock \emph{Advances in neural information processing systems}, 33:\penalty0 6840--6851, 2020.

\bibitem[Huang et~al.(2022)Huang, Tang, Dong, and Xu]{huang2022draw}
Huang, N., Tang, F., Dong, W., and Xu, C.
\newblock Draw your art dream: Diverse digital art synthesis with multimodal guided diffusion.
\newblock In \emph{Proceedings of the 30th ACM International Conference on Multimedia}, pp.\  1085--1094, 2022.

\bibitem[Ku et~al.(2024)Ku, Li, Zhang, Lu, Fu, Zhuang, and Chen]{kuimagenhub}
Ku, M., Li, T., Zhang, K., Lu, Y., Fu, X., Zhuang, W., and Chen, W.
\newblock Imagenhub: Standardizing the evaluation of conditional image generation models.
\newblock In \emph{The Twelfth International Conference on Learning Representations}, 2024.

\bibitem[Langdon \& Harman(2014)Langdon and Harman]{langdon2014optimizing}
Langdon, W.~B. and Harman, M.
\newblock Optimizing existing software with genetic programming.
\newblock \emph{IEEE Transactions on Evolutionary Computation}, 19\penalty0 (1):\penalty0 118--135, 2014.

\bibitem[Langdon et~al.(2015)Langdon, Lam, Petke, and Harman]{langdon2015improving}
Langdon, W.~B., Lam, B. Y.~H., Petke, J., and Harman, M.
\newblock Improving cuda dna analysis software with genetic programming.
\newblock In \emph{Proceedings of the 2015 Annual Conference on Genetic and Evolutionary Computation}, pp.\  1063--1070, 2015.

\bibitem[Lee et~al.(2023)Lee, Liu, Ryu, Watkins, Du, Boutilier, Abbeel, Ghavamzadeh, and Gu]{lee2023aligning}
Lee, K., Liu, H., Ryu, M., Watkins, O., Du, Y., Boutilier, C., Abbeel, P., Ghavamzadeh, M., and Gu, S.~S.
\newblock Aligning text-to-image models using human feedback.
\newblock \emph{arXiv preprint arXiv:2302.12192}, 2023.

\bibitem[Liu \& Chilton(2022)Liu and Chilton]{liu2022design}
Liu, V. and Chilton, L.~B.
\newblock Design guidelines for prompt engineering text-to-image generative models.
\newblock In \emph{Proceedings of the 2022 CHI conference on human factors in computing systems}, pp.\  1--23, 2022.

\bibitem[Martins et~al.(2023)Martins, Cunha, Correia, and Machado]{martins2023towards}
Martins, T., Cunha, J.~M., Correia, J., and Machado, P.
\newblock Towards the evolution of prompts with metaprompter.
\newblock In \emph{International Conference on Computational Intelligence in Music, Sound, Art and Design (Part of EvoStar)}, pp.\  180--195. Springer, 2023.

\bibitem[Peer et~al.(2017)Peer, Brandimarte, Samat, and Acquisti]{peer2017beyond}
Peer, E., Brandimarte, L., Samat, S., and Acquisti, A.
\newblock Beyond the turk: Alternative platforms for crowdsourcing behavioral research.
\newblock \emph{Journal of experimental social psychology}, 70:\penalty0 153--163, 2017.

\bibitem[Petke et~al.(2017)Petke, Haraldsson, Harman, Langdon, White, and Woodward]{petke2017genetic}
Petke, J., Haraldsson, S.~O., Harman, M., Langdon, W.~B., White, D.~R., and Woodward, J.~R.
\newblock Genetic improvement of software: a comprehensive survey.
\newblock \emph{IEEE Transactions on Evolutionary Computation}, 22\penalty0 (3):\penalty0 415--432, 2017.

\bibitem[Ramesh et~al.(2021)Ramesh, Pavlov, Goh, Gray, Voss, Radford, Chen, and Sutskever]{ramesh2021zero}
Ramesh, A., Pavlov, M., Goh, G., Gray, S., Voss, C., Radford, A., Chen, M., and Sutskever, I.
\newblock Zero-shot text-to-image generation.
\newblock In \emph{International conference on machine learning}, pp.\  8821--8831. Pmlr, 2021.

\bibitem[Redmon(2016)]{redmon2016you}
Redmon, J.
\newblock You only look once: Unified, real-time object detection.
\newblock In \emph{Proceedings of the IEEE conference on computer vision and pattern recognition}, 2016.

\bibitem[Reed et~al.(2016)Reed, Akata, Yan, Logeswaran, Schiele, and Lee]{reed2016generative}
Reed, S., Akata, Z., Yan, X., Logeswaran, L., Schiele, B., and Lee, H.
\newblock Generative adversarial text to image synthesis.
\newblock In \emph{International conference on machine learning}, pp.\  1060--1069. PMLR, 2016.

\bibitem[Rombach et~al.(2022)Rombach, Blattmann, Lorenz, Esser, and Ommer]{rombach2022high}
Rombach, R., Blattmann, A., Lorenz, D., Esser, P., and Ommer, B.
\newblock High-resolution image synthesis with latent diffusion models.
\newblock In \emph{Proceedings of the IEEE/CVF conference on computer vision and pattern recognition}, pp.\  10684--10695, 2022.

\bibitem[Saharia et~al.(2022)Saharia, Chan, Saxena, Li, Whang, Denton, Ghasemipour, Gontijo~Lopes, Karagol~Ayan, Salimans, et~al.]{saharia2022photorealistic}
Saharia, C., Chan, W., Saxena, S., Li, L., Whang, J., Denton, E.~L., Ghasemipour, K., Gontijo~Lopes, R., Karagol~Ayan, B., Salimans, T., et~al.
\newblock Photorealistic text-to-image diffusion models with deep language understanding.
\newblock \emph{Advances in neural information processing systems}, 35:\penalty0 36479--36494, 2022.

\bibitem[Santana(2022)]{santana2022}
Santana, G.
\newblock Stable-diffusion-prompts.
\newblock \url{https://huggingface.co/datasets/Gustavosta/Stable-Diffusion-Prompts/blob/main/data/train.parquet}, 2022.
\newblock Accessed: November 10, 2024.

\bibitem[Tao et~al.(2022)Tao, Tang, Wu, Jing, Bao, and Xu]{tao2022df}
Tao, M., Tang, H., Wu, F., Jing, X.-Y., Bao, B.-K., and Xu, C.
\newblock Df-gan: A simple and effective baseline for text-to-image synthesis.
\newblock In \emph{Proceedings of the IEEE/CVF conference on computer vision and pattern recognition}, pp.\  16515--16525, 2022.

\bibitem[Wang et~al.(2024)Wang, Liu, Hsieh, and Gong]{wang2024discrete}
Wang, R., Liu, T., Hsieh, C.-J., and Gong, B.
\newblock On discrete prompt optimization for diffusion models.
\newblock \emph{arXiv preprint arXiv:2407.01606}, 2024.

\bibitem[Wang et~al.(2022)Wang, Montoya, Munechika, Yang, Hoover, and Chau]{wang2022diffusiondb}
Wang, Z.~J., Montoya, E., Munechika, D., Yang, H., Hoover, B., and Chau, D.~H.
\newblock Diffusiondb: A large-scale prompt gallery dataset for text-to-image generative models.
\newblock \emph{arXiv preprint arXiv:2210.14896}, 2022.

\bibitem[Wilcoxon(1992)]{wilcoxon1992individual}
Wilcoxon, F.
\newblock Individual comparisons by ranking methods.
\newblock In \emph{Breakthroughs in statistics: Methodology and distribution}, pp.\  196--202. Springer, 1992.

\bibitem[Wu et~al.(2023)Wu, Sun, Zhu, Zhao, and Li]{wu2023human}
Wu, X., Sun, K., Zhu, F., Zhao, R., and Li, H.
\newblock Human preference score: Better aligning text-to-image models with human preference.
\newblock In \emph{Proceedings of the IEEE/CVF International Conference on Computer Vision}, pp.\  2096--2105, 2023.

\bibitem[Xu et~al.(2024)Xu, Liu, Wu, Tong, Li, Ding, Tang, and Dong]{xu2024imagereward}
Xu, J., Liu, X., Wu, Y., Tong, Y., Li, Q., Ding, M., Tang, J., and Dong, Y.
\newblock Imagereward: Learning and evaluating human preferences for text-to-image generation.
\newblock \emph{Advances in Neural Information Processing Systems}, 36, 2024.

\bibitem[Ye et~al.(2023)Ye, Zhang, Liu, Han, and Yang]{ye2023ip}
Ye, H., Zhang, J., Liu, S., Han, X., and Yang, W.
\newblock {IP}-{A}dapter: Text compatible image prompt adapter for text-to-image diffusion models.
\newblock \emph{arXiv preprint arXiv:2308.06721}, 2023.

\bibitem[Yip(2023)]{yip2023}
Yip, E.
\newblock 100+ negative prompts everyone are using.
\newblock \url{https://medium.com/stablediffusion/100-negative-prompts-everyone-are-using-c71d0ba33980}, 2023.
\newblock Accessed: November 10, 2024.

\bibitem[Yuan \& Banzhaf(2020)Yuan and Banzhaf]{yuan2020toward}
Yuan, Y. and Banzhaf, W.
\newblock Toward better evolutionary program repair: An integrated approach.
\newblock \emph{ACM Transactions on Software Engineering and Methodology (TOSEM)}, 29\penalty0 (1):\penalty0 1--53, 2020.

\bibitem[Zhang et~al.(2023)Zhang, Rao, and Agrawala]{zhang2023adding}
Zhang, L., Rao, A., and Agrawala, M.
\newblock Adding conditional control to text-to-image diffusion models.
\newblock In \emph{Proceedings of the IEEE/CVF International Conference on Computer Vision}, pp.\  3836--3847, 2023.

\end{thebibliography}
\bibliographystyle{icml2024}

\newpage
\appendix
\onecolumn

\section{Workflow and Mutation Operator Settings}
\label{app:appendix_workflow_mutation_settings}

In our experiments, we use a text-to-image generation workflow using the modules: \textit{Empty Latent Image}, \textit{Load Checkpoint}, \textit{CLIP Text Encode (Prompt)} for the positive and the negative prompt, \textit{KSampler}, \textit{VAE Decode}, and \textit{Save Image}. In the \textit{Empty Latent Image} module, we set the image dimensions to 512x512 as default image size for all generated images. The default settings for the \textit{KSampler} module as well as the possible values and ranges for the \texttt{ksampler} mutation operator are presented in Table~\ref{tab:ksampler_settings}.  

\begin{table*}[!h]
    \centering
    \caption{Default property values and possible values / ranges of the \texttt{ksampler} mutation operator.}
    \small
    \begin{tabular}{l c c}
    \toprule
    \textbf{Property} & \textbf{Default} & \textbf{Possible Values / Ranges}  \\
    \midrule
    \vspace{1mm}
    seed          & 1337\textbf{*} & $[0, 100000]$ \\
    \vspace{1mm}
    steps         & 20     & $[1,200)$ \\
    \vspace{1mm}
    cfg           & 8.0    & $[0.0, 25.0)$ \\
                  &        & euler, euler\_ancestral, heun, heunpp2, dpm\_2, dpm\_2\_ancestral, lms, dpm\_fast, dpm\_adaptive,  \\
    sampler\_name & dpm\_2 & dpmpp\_2s\_ancestral, dpmpp\_sde, dpmpp\_sde\_gpu, dpmpp\_2m, dpmpp\_2m\_sde, dpmpp\_2m\_sde\_gpu, \\
    \vspace{1mm}
                  &        & dpmpp\_3m\_sde, dpmpp\_3m\_sde\_gpu, ddpm, lcm, ddim, uni\_pc, uni\_pc\_bh2 \\
    \vspace{1mm}
    scheduler     & normal & normal, karras, exponential, sgm\_uniform, simple, ddim\_uniform \\
    denoise       & $1.0$  & $[0.00, 1.00]$ \\
    \bottomrule
    \end{tabular}
    \parbox{\linewidth}{
        \vspace{0.5mm}
        \footnotesize \;\;\;\textbf{*}The default seed was replaced in each of the $10$ experimental runs.
    }
    \label{tab:ksampler_settings}
\end{table*}

The \texttt{prompt\_llm} mutation operator requests an LLM to optimize the workflow's prompt for image generation, which in turn requires prompts. The prompts we used for this are presented in Figs.~\ref{fig:positive_prompt} (positive prompt) and~\ref{fig:negative_prompt} (negative prompt). The placeholder \texttt{[PROMPT]} shown in the figures is replaced by the image generation prompt to be improved. For the improvement of the negative prompts, it may seem obvious to add the corresponding positive prompt to the LLM request, but in our tests this did not lead to any improvement in the prompt quality. 

\begin{figure}[h!]
    \centering
    \begin{minipage}{0.95\textwidth}
        \begin{lstlisting}[basicstyle=\ttfamily]
Rewrite the following positive prompt such that it works best for a diffusion 
model for text to image generation: "[PROMPT]". Give a short description 
followed by a few comma (,) separated short image feature descriptions. 
Return only the updated prompt and nothing else.
        \end{lstlisting}
    \end{minipage}
    \caption{Prompt used by the \texttt{prompt\_llm} mutation operator to improve the workflow's positive prompt.}
    \label{fig:positive_prompt}
\end{figure}

\begin{figure}[h!]
    \centering
    \begin{minipage}{0.95\textwidth}
        \begin{lstlisting}[basicstyle=\ttfamily]
Replace the following negative prompt with a new one such that it works best 
for a diffusion model for text to image generation: "[PROMPT]". Return a 
comma (,) separated list for the new prompt. Return only the updated prompt 
and nothing else.
        \end{lstlisting}
    \end{minipage}
    \caption{Prompt used by the \texttt{prompt\_llm} mutation operator to improve the workflow's negative prompt.}
    \label{fig:negative_prompt}
\end{figure}

To give a complete overview, Fig.~\ref{fig:json_workflow} shows the text-to-image workflow used in the experiments with default values and prompts in JSON format. Please note that we randomly replaced the seed and the default checkpoint model in each of the $10$ runs to ensure a fair evaluation. Furthermore, the workflow's positive prompt was replaced by the appropriate benchmark prompt for the experiments.

\clearpage

\section{Details of the Models Used in the Experiments}
\label{app:model_versions}

ComfyGI uses machine learning models in various places. Table~\ref{tab:llm_versions} shows the LLMs (with version information) used by the \texttt{prompt\_llm} mutation operator. Table~\ref{tab:imggen_models} shows the image generation models used by the workflows and the \texttt{checkpoint} mutation operator. In addition to the model names, the table shows also the links to the used \texttt{*.safetensors} files.

\begin{table*}[!h]
    \centering
    \caption{LLM models used by the \texttt{prompt\_llm} mutation operator.}
    \small
    \begin{tabular}{l l c}
    \toprule
    \textbf{Model} & \textbf{Link} & \textbf{Version / Hash}  \\
    \midrule
    llama3.1:8b      & \url{https://ollama.com/library/llama3.1:8b}      & 91ab477bec9d \\
    gemma2:9b        & \url{https://ollama.com/library/gemma2:9b}        & ff02c3702f32 \\
    mistral-nemo:12b & \url{https://ollama.com/library/mistral-nemo:12b} & 994f3b8b7801 \\
    \bottomrule
    \end{tabular}
    \label{tab:llm_versions}
\end{table*}

\begin{table*}[!h]
    \centering
    \caption{Image generation models used for the workflows / by the \texttt{checkpoint} mutation operator. All models were downloaded on August 12, 2024.}
    \small
    \begin{tabular}{l p{12.4cm}}
    \toprule
    \textbf{Model} & \textbf{Link} \\
    \midrule
    Stable Diffusion 1.5          & \url{https://huggingface.co/runwayml/stable-diffusion-v1-5/resolve/main/v1-5-pruned.safetensors} \\
    Stable Diffusion 2            & \url{https://huggingface.co/stabilityai/stable-diffusion-2/resolve/main/768-v-ema.safetensors} \\
    Stable Diffusion 3 Medium     & \url{https://huggingface.co/ckpt/stable-diffusion-3-medium/resolve/main/sd3_medium_incl_clips.safetensors} \\
    Stable Diffusion XL Turbo 1.0 & \url{https://huggingface.co/stabilityai/sdxl-turbo/resolve/main/sd_xl_turbo_1.0_fp16.safetensors} \\
    Stable Diffusion XL Base 1.0  & \url{https://huggingface.co/stabilityai/stable-diffusion-xl-base-1.0/resolve/main/sd_xl_base_1.0.safetensors} \\
    Dreamlike Photoreal 2.0       & \url{https://huggingface.co/dreamlike-art/dreamlike-photoreal-2.0/resolve/main/dreamlike-photoreal-2.0.safetensors} \\
    DreamShaper 3.3               & \url{https://huggingface.co/Lykon/DreamShaper/resolve/main/DreamShaper_3.3.safetensors} \\
    Realistic Vision 6.0          & \url{https://huggingface.co/SG161222/Realistic_Vision_V6.0_B1_noVAE/resolve/main/Realistic_Vision_V6.0_NV_B1.safetensors} \\
    ReV Animated 1.2.2            & \url{https://huggingface.co/danbrown/RevAnimated-v1-2-2/resolve/main/rev-animated-v1-2-2.safetensors} \\
    \bottomrule
    \end{tabular}
    \label{tab:imggen_models}
\end{table*}

\begin{figure}[h!]
    \centering
    \begin{minipage}{0.95\textwidth}
        \tiny
\begin{lstlisting}[language=json]
{
  "3": {
    "inputs": {
      "seed": 1337,
      "steps": 20,
      "cfg": 8,
      "sampler_name": "dpm_2",
      "scheduler": "normal",
      "denoise": 1,
      "model": ["4", 0],
      "positive": ["6", 0],
      "negative": ["7", 0],
      "latent_image": ["5", 0]
    },
    "class_type": "KSampler",
    "_meta": {
      "title": "KSampler"
    }
  },
  "4": {
    "inputs": {
      "ckpt_name": "sd3_medium_incl_clips.safetensors"
    },
    "class_type": "CheckpointLoaderSimple",
    "_meta": {
      "title": "Load Checkpoint"
    }
  },
  "5": {
    "inputs": {
      "width": 512,
      "height": 512,
      "batch_size": 1
    },
    "class_type": "EmptyLatentImage",
    "_meta": {
      "title": "Empty Latent Image"
    }
  },
  "6": {
    "inputs": {
      "text": "An astronaut on a horse, detailed, 4k, high quality",
      "clip": ["4", 1]
    },
    "class_type": "CLIPTextEncode",
    "_meta": {
      "title": "CLIP Text Encode (Prompt)"
    }
  },
  "7": {
    "inputs": {
      "text": "worse, bad quality",
      "clip": ["4", 1]
    },
    "class_type": "CLIPTextEncode",
    "_meta": {
      "title": "CLIP Text Encode (Prompt)"
    }
  },
  "8": {
    "inputs": {
      "samples": ["3", 0],
      "vae": ["4", 2]
    },
    "class_type": "VAEDecode",
    "_meta": {
      "title": "VAE Decode"
    }
  },
  "9": {
    "inputs": {
      "filename_prefix": "ComfyUI",
      "images": ["8", 0]
    },
    "class_type": "SaveImage",
    "_meta": {
      "title": "Save Image"
    }
  }
}
\end{lstlisting} 
    \end{minipage}
    \caption{The text-to-image workflow used in the experiments with default values and prompts. Please note that we randomly replaced the seed and the default checkpoint model in each of the $10$ runs to ensure a fair evaluation. Furthermore, the positive prompt was replaced by the appropriate benchmark prompt for the experiments.}
    \label{fig:json_workflow}
\end{figure}

\section{Benchmark Prompts}
\label{app:benchmark_problems}

To evaluate the performance of ComfyGI, we randomly sampled $42$ prompts from $14$ categories from the ImagenHub benchmark suite \cite{kuimagenhub}. Table~\ref{tab:benchmark_prompts} shows the prompts used and the corresponding categories.

\begin{table*}[!h]
    \centering
    \caption{Prompts used for the experiments with their corresponding categories.}
    \small
    \begin{tabular}{l p{13.9cm}}
    \toprule
    \textbf{Category} & \textbf{Prompt} \\
    \midrule
    Colors & an orange colored sandwich \\
    Colors & a blue cup and a green cell phone \\
    Colors & a red colored car \\
    Conflicting & an elephant under the sea \\
    Conflicting & a horse riding an astronaut \\
    Conflicting & a panda making latte art \\
    Counting & four dogs on the street \\
    Counting & two cars on the street \\
    Counting & one cat and two dogs sitting on the grass \\
    DALL-E & a couple of glasses are sitting on a table \\
    DALL-E & a triangular orange picture frame. an orange picture frame in the shape of a triangle \\
    DALL-E & an emoji of a baby panda wearing a red hat, green gloves, red shirt, and green pants \\
    Descriptions & a mechanical or electrical device for measuring time \\
    Descriptions & a large motor vehicle carrying passengers by road, typically one serving the public on a fixed route and for a fare \\
    Descriptions & an american multinational technology company that focuses on artificial intelligence, search engine, online advertising, cloud computing, computer software, quantum computing, e-commerce, and consumer electronics \\
    Gary Marcus et al.  & a donkey and an octopus are playing a game. the donkey is holding a rope on one end, the octopus is holding onto the other. the donkey holds the rope in its mouth. a cat is jumping over the rope \\
    Gary Marcus et al.  & paying for a quarter-sized pizza with a pizza-sized quarter \\
    Gary Marcus et al.  & a tomato has been put on top of a pumpkin on a kitchen stool. there is a fork sticking into the pumpkin. the scene is viewed from above \\
    Hard & a person wearing a black suit jacket, collared shirt, blue patterned tie and a pocket square \\
    Hard & a silver , orange , and grey   train 's closed doors  \\
    Hard & close together group of white, beige, and brown sheep, from the rear, standing in grass at fence gate \\
    Misc & beautiful gorgeous elegant porcelain ivory fair skin mechanoid woman, close - up, sharp focus, studio light, iris van herpen haute couture headdress made of rhizomorphs, daisies, arches, brackets, herbs, colorful corals, fractal mushrooms, puffballs, octane render, ultra sharp, 8 k \\
    Misc & photo cartoon bd illustration comic manga painting of cloakroom environement : 5 fantasy environement, digital painting : 1 fat brush concept sketch artist bd enki bilal : 1 0 \\
    Misc & a photograph of a futuristic street scene, brutalist style architecture, straight edges, shop signs, video screens, finely detailed oil painting, impasto brush strokes, soft light, 8 k, dramatic composition, dramatic lighting, sharp focus, octane render, masterpiece, by adrian ghenie and jenny saville and zhang jingna \\
    Misspellings & an instqrumemnt used for cutting cloth, paper, axdz othr thdin mteroial, consamistng of two blades lad one on tvopb of the other and fhastned in tle mixdqdjle so as to bllow them txo be pened and closed by thumb and fitngesr inserted tgrough rings on kthe end oc thei vatndlzes \\
    Misspellings & tcennis rpacket \\
    Misspellings & a smafml vessef epropoeilled on watvewr by ors, sauls, or han engie \\
    Positional & a tennis racket underneath a traffic light \\
    Positional & a wine glass on top of a dog \\
    Positional & a pizza on the right of a suitcase \\
    Rare Words & artophagous \\
    Rare Words & matutinal \\
    Rare Words & backlotter \\
    Realism & hyperrealism backstreet, in front of a massive building, cinematic, neon signs, high resolution, realistic lighting, octane render, hyper realistic, 8k \\
    Realism & image of a huge black castle in the sky surrounded by black vines \\
    Realism & image of a fashion model standing in an audience, wearing black balenciaga hooded jacket, looking away from camera, covered in plastic, 3d, 4k, fashion shoot \\
    Reddit & a maglev train going vertically downward in high speed, new york times photojournalism \\
    Reddit & mcdonalds church \\
    Reddit & an ancient egyptian painting depicting an argument over whose turn it is to take out the trash \\
    Text & a sign that says 'diffusion' \\
    Text & a storefront with 'diffusion' written on it \\
    Text & new york skyline with 'diffusion' written with fireworks on the sky \\
    \bottomrule
    \end{tabular}
    \label{tab:benchmark_prompts}
\end{table*}

\section{Details on the Human Evaluation Method}
\label{app:human_eval}

In our study, we also conducted a human evaluation of the results of ComfyGI with $100$ participants. Before the participants started the survey, we instructed them on what they should pay attention to in the study:

\begin{itemize}
    \item \textbf{Alignment} with the textual description (How well does the image capture the given text?)
    \item \textbf{Quality} of the image (How appealing is the image in your personal opinion?)
\end{itemize}

\begin{figure}[ht!]
    \centering
    \subfloat[\textbf{Example A}: Both images are aligned with the description. Therefore, you should select the image that is more appealing to you.]{
        \includegraphics[width=0.7\columnwidth]{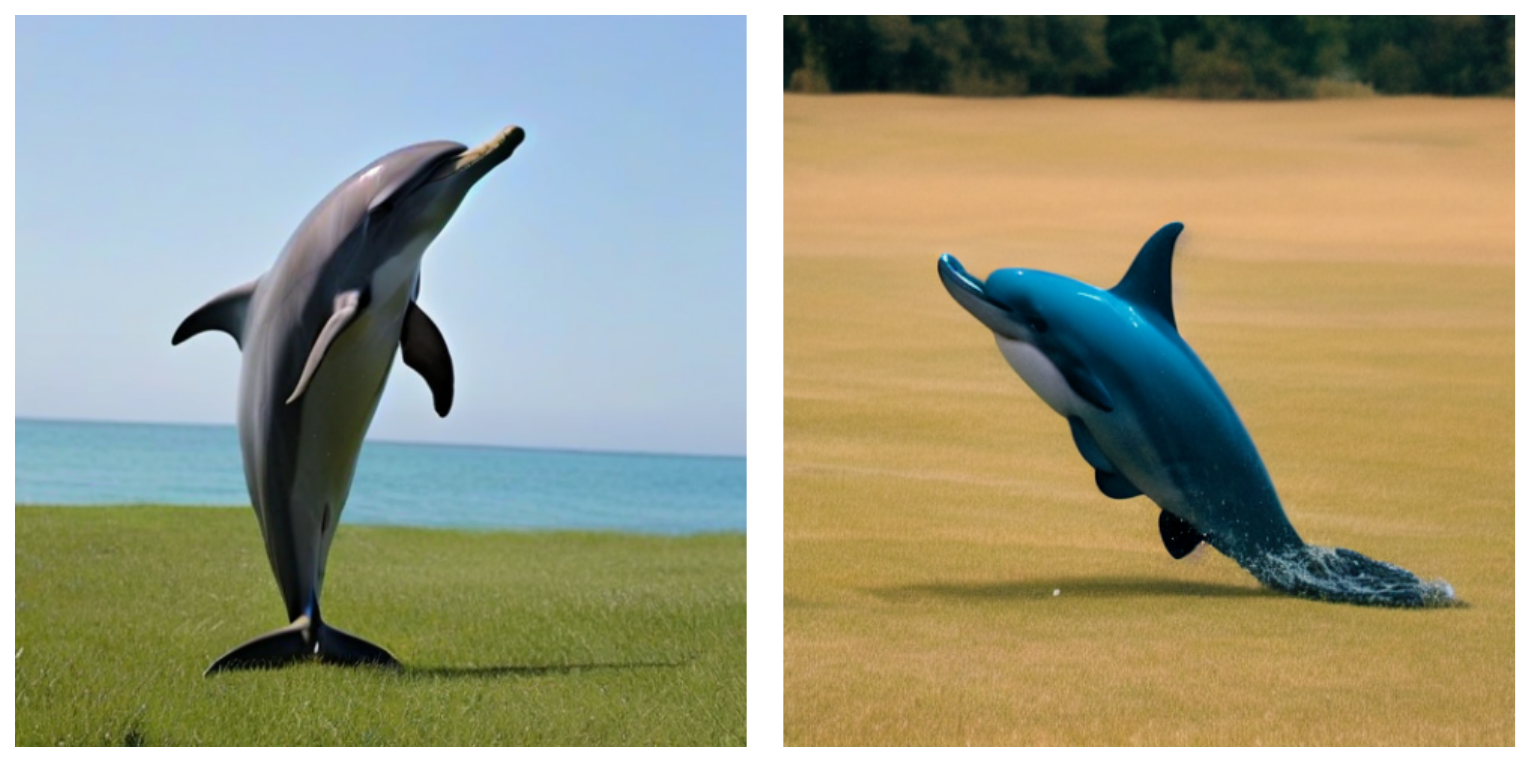}
        \label{fig:priming_1}
    }\hfill
    \subfloat[\textbf{Example B}: The second image might be more appealing to you, but the first image is aligned with the description, while the second image is not (a goldfish is displayed instead of a dolphin). Therefore, you should select the first image with the dolphin.]{
        \includegraphics[width=0.7\columnwidth]{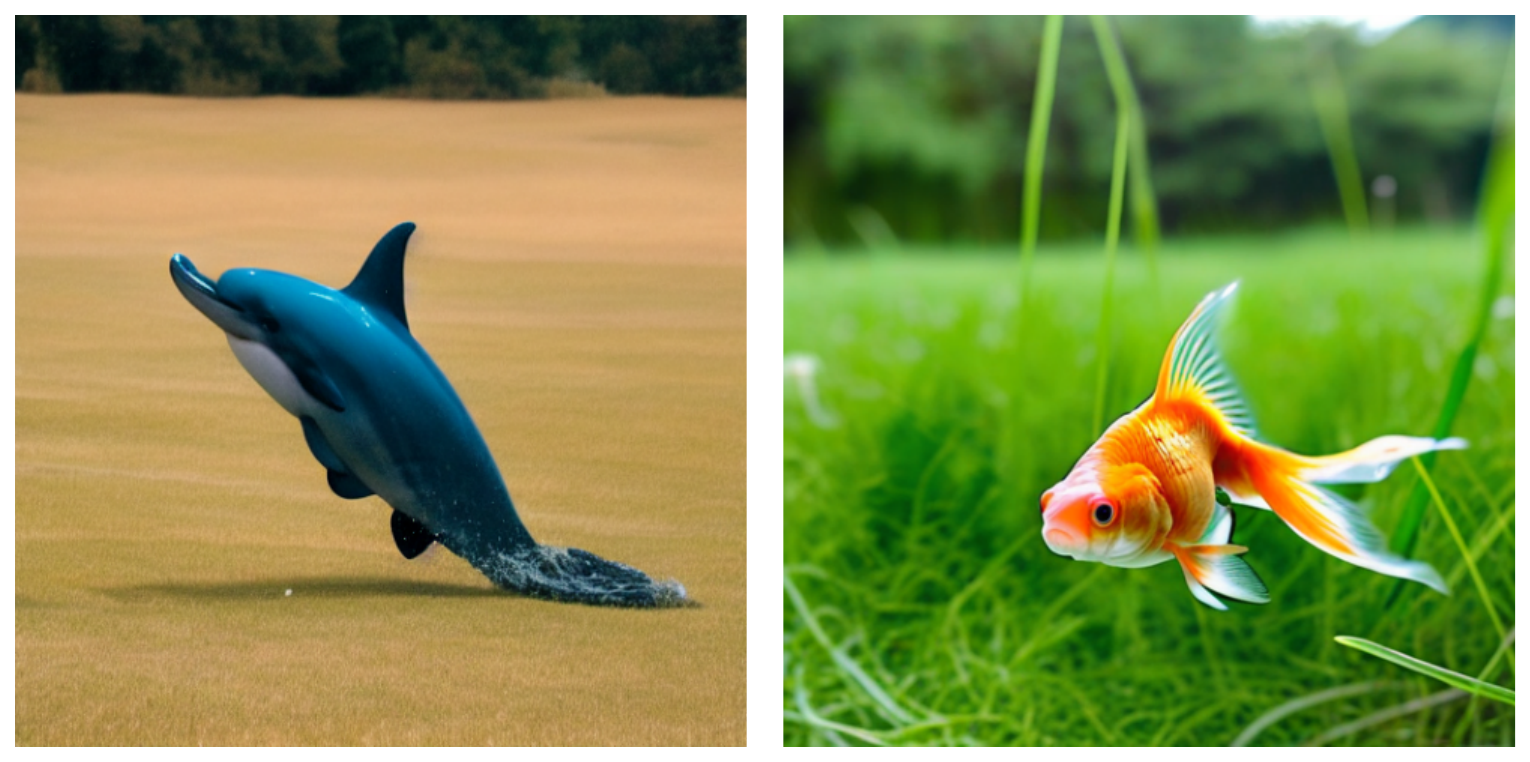}
        \label{fig:priming_2}
    }\hfill
    \subfloat[\textbf{Example C}: Both images are not aligned with the description. Therefore, you should select the image you find more appealing.]{
        \includegraphics[width=0.7\columnwidth]{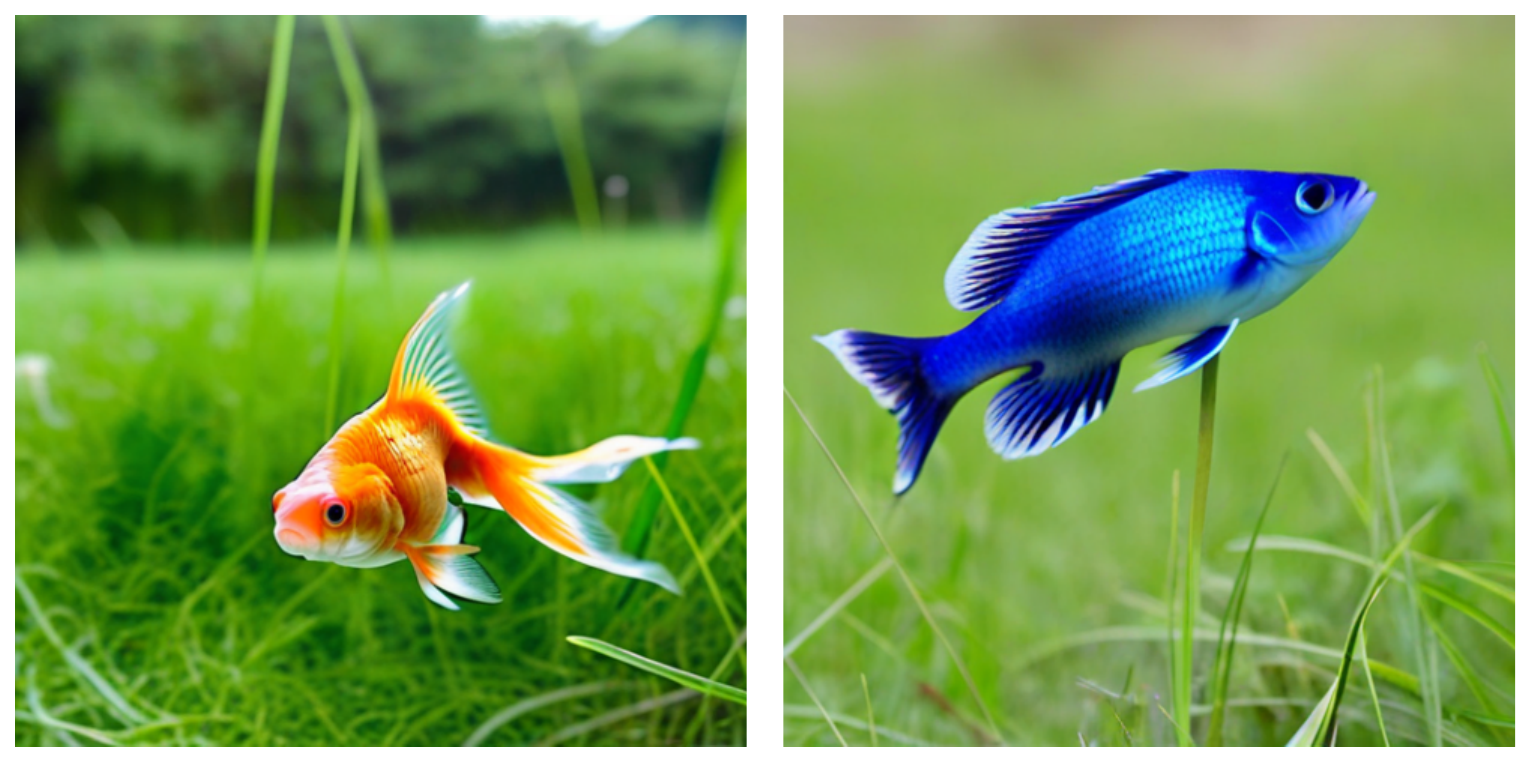}
        \label{fig:priming_3}
    }
    \caption{Examples used for priming the participants.}
    \label{fig:priming_examples}
\end{figure}

We also told them, that an image that is aligned with the prompt should be preferred, even if it is of lower quality than the unaligned image. Further we showed the participants some example image pairs. These examples are shown in Fig.~\ref{fig:priming_examples}. The descriptions of the example image pairs as shown to the participants are shown in sub-captions \ref{fig:priming_1}-\ref{fig:priming_3}.

\begin{figure}[ht!]
    \centering
    \subfloat[Please select the image that best aligns with the description and is most appealing to you. The description is as follows: \textbf{a brown dog in a green garden with red flowers}.]{
        \includegraphics[width=0.7\columnwidth]{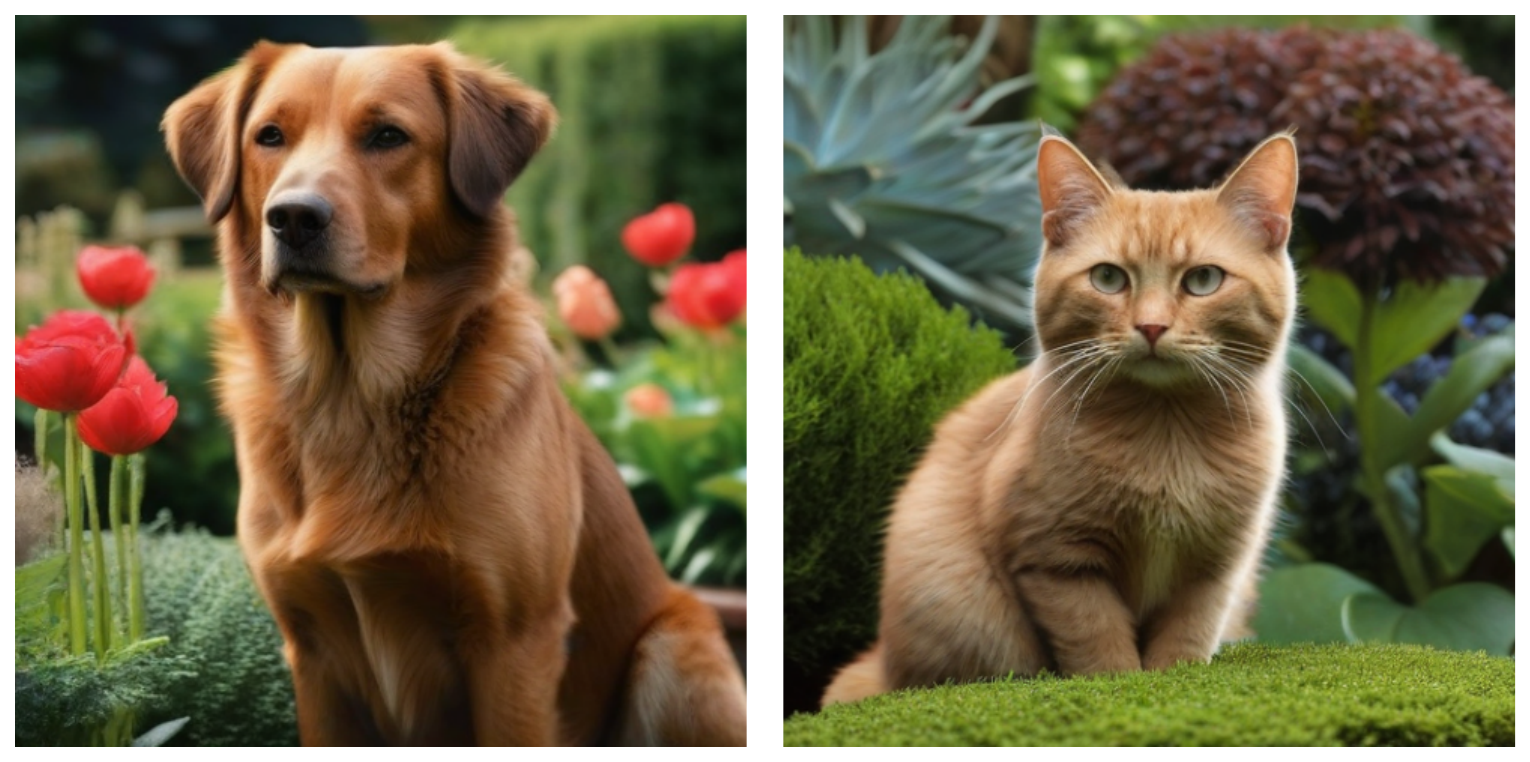}
        \label{fig:att_check_1}
    }\hfill
    \subfloat[Please select the image that best aligns with the description and is most appealing to you. The description is as follows: \textbf{a big red car on a street in a city}.]{
        \includegraphics[width=0.7\columnwidth]{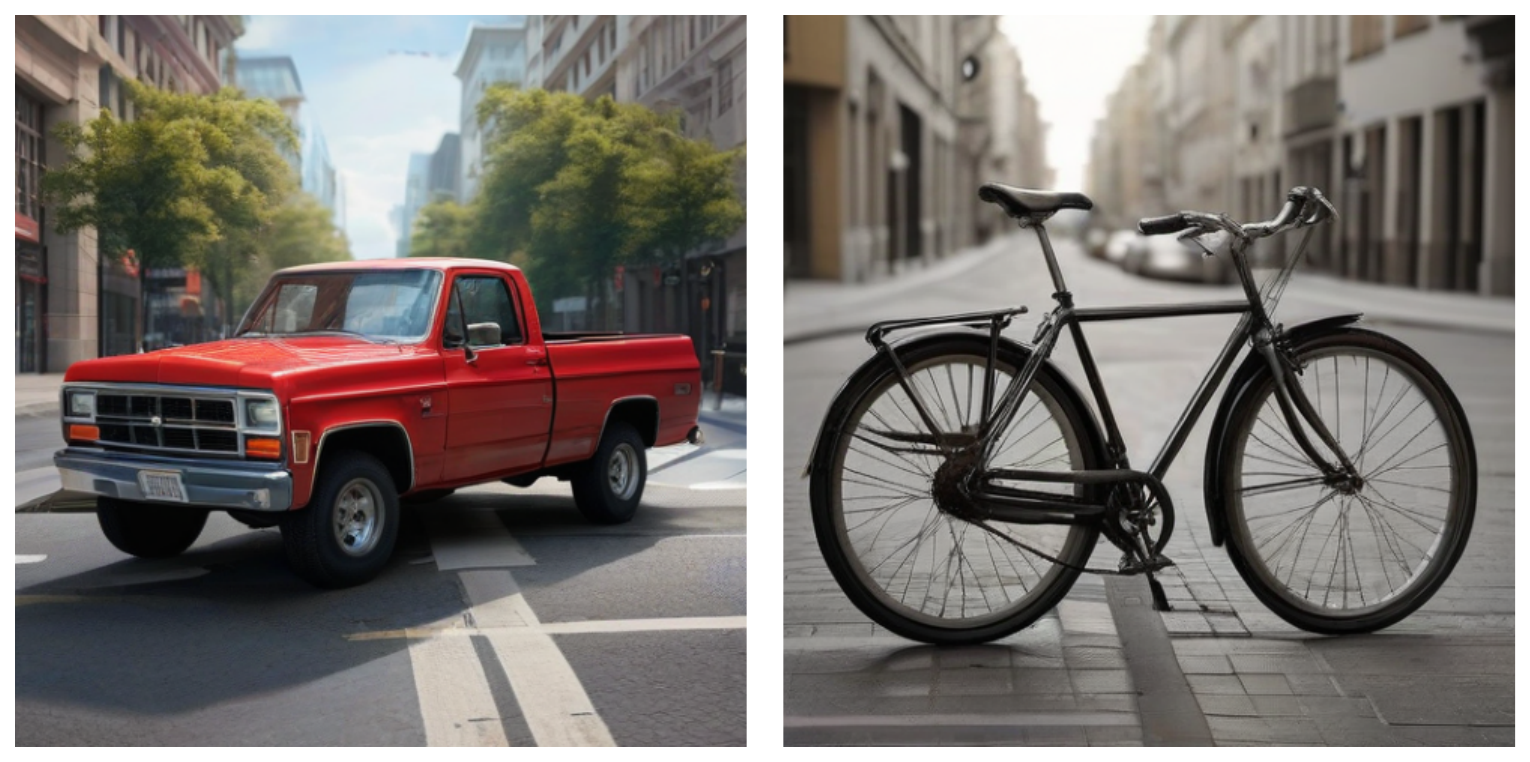}
        \label{fig:att_check_2}
    }\hfill
    \subfloat[Please select the image that best aligns with the description and is most appealing to you. The description is as follows: \textbf{a wooden table on a gray carpet in front of a couch}.]{
        \includegraphics[width=0.7\columnwidth]{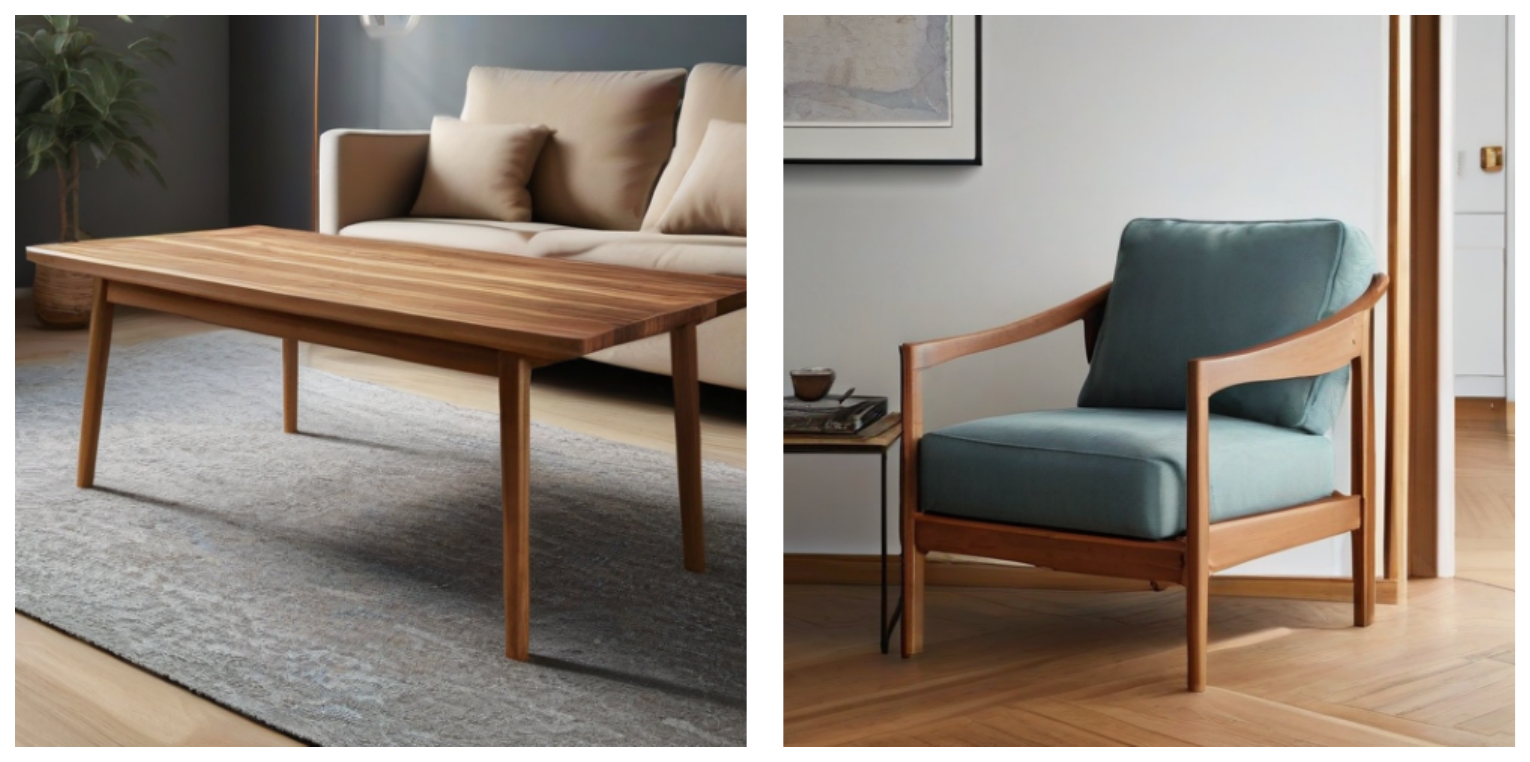}
        \label{fig:att_check_3}
    }
    \caption{Attention checks.}
    \label{fig:att_checks}
\end{figure}

In addition to that, we also integrated three image pairs as attention checks into the survey. These image pairs are shown in Fig.~\ref{fig:att_checks} with their corresponding descriptions shown to the participants in the sub-captions. 

\begin{figure*}[!h]
    \centering
    \includegraphics[width=\textwidth]{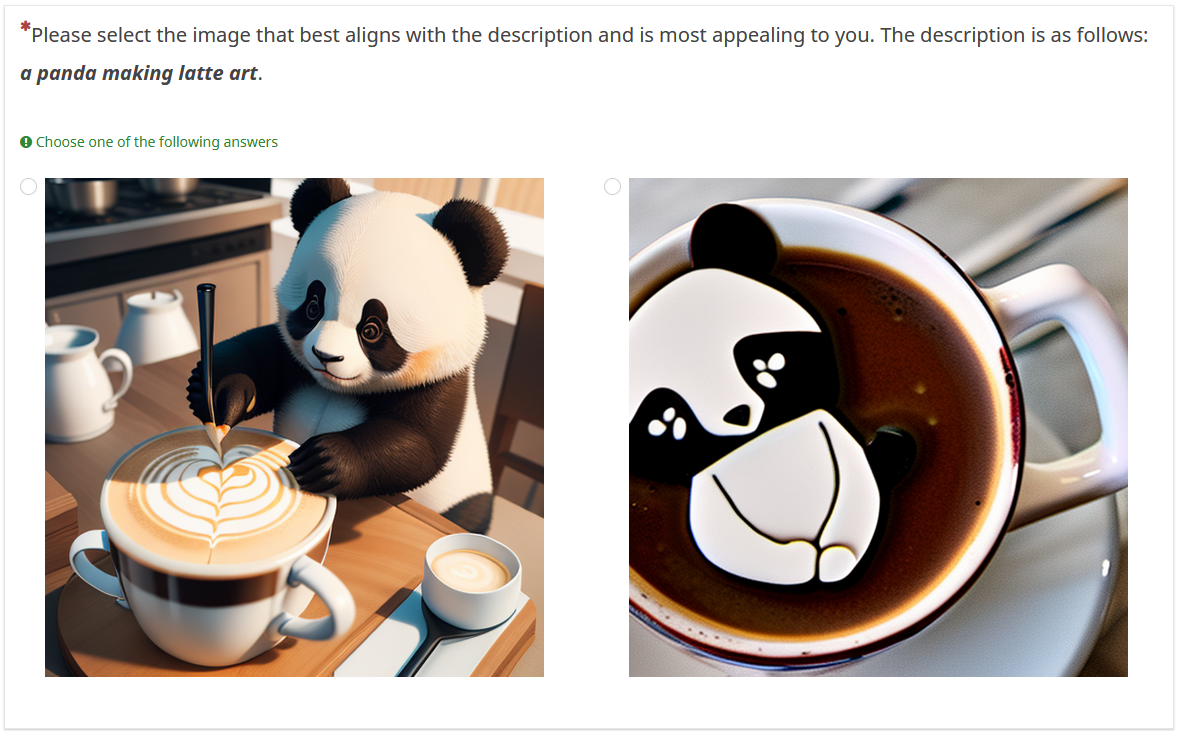}
    \caption{Example of a question displayed to human annotators. The position (left or right) of the initial and optimized image is randomized for each participant to avoid positional bias.}
    \label{fig:questionnaire_example}
\end{figure*}

A standard example question (image pair) is given in Fig.~\ref{fig:questionnaire_example} exactly as shown to the participants.

\clearpage
\section{Intermediary Steps of the Image Examples}
\label{app:images_intermediary}

\begin{figure*}[!h]
    \centering
    \includegraphics[width=\textwidth]{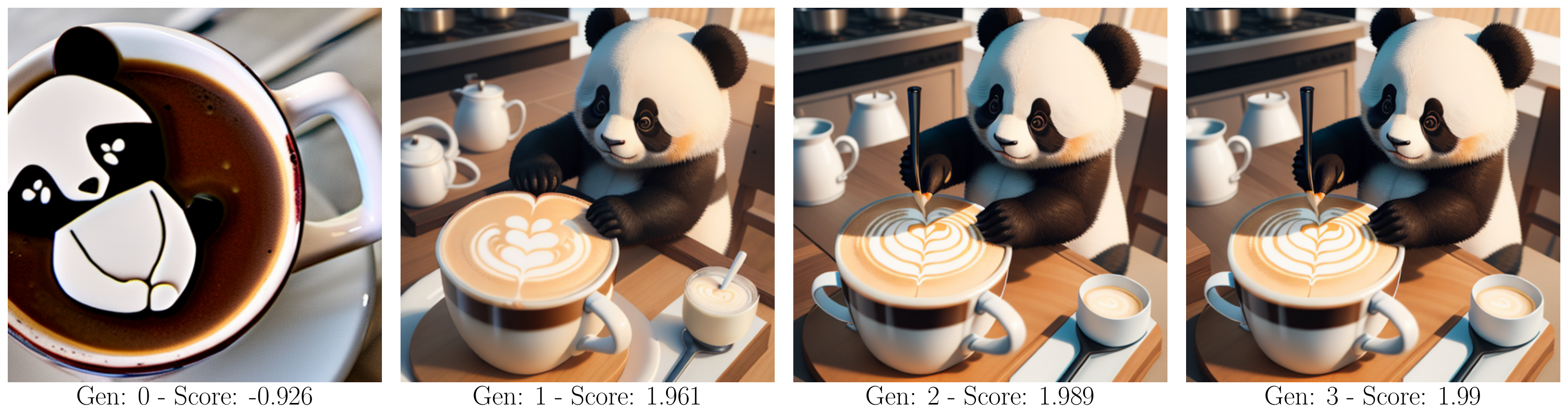}
    \vspace{-8mm} 
    \caption{An example for image improvement with ComfyGI over several generations for the prompt \textit{``a panda making latte art''}.}
    \label{fig:panda_latte_art_evolution}
\end{figure*}

\begin{figure*}[!h]
    \centering
    \includegraphics[width=\textwidth]{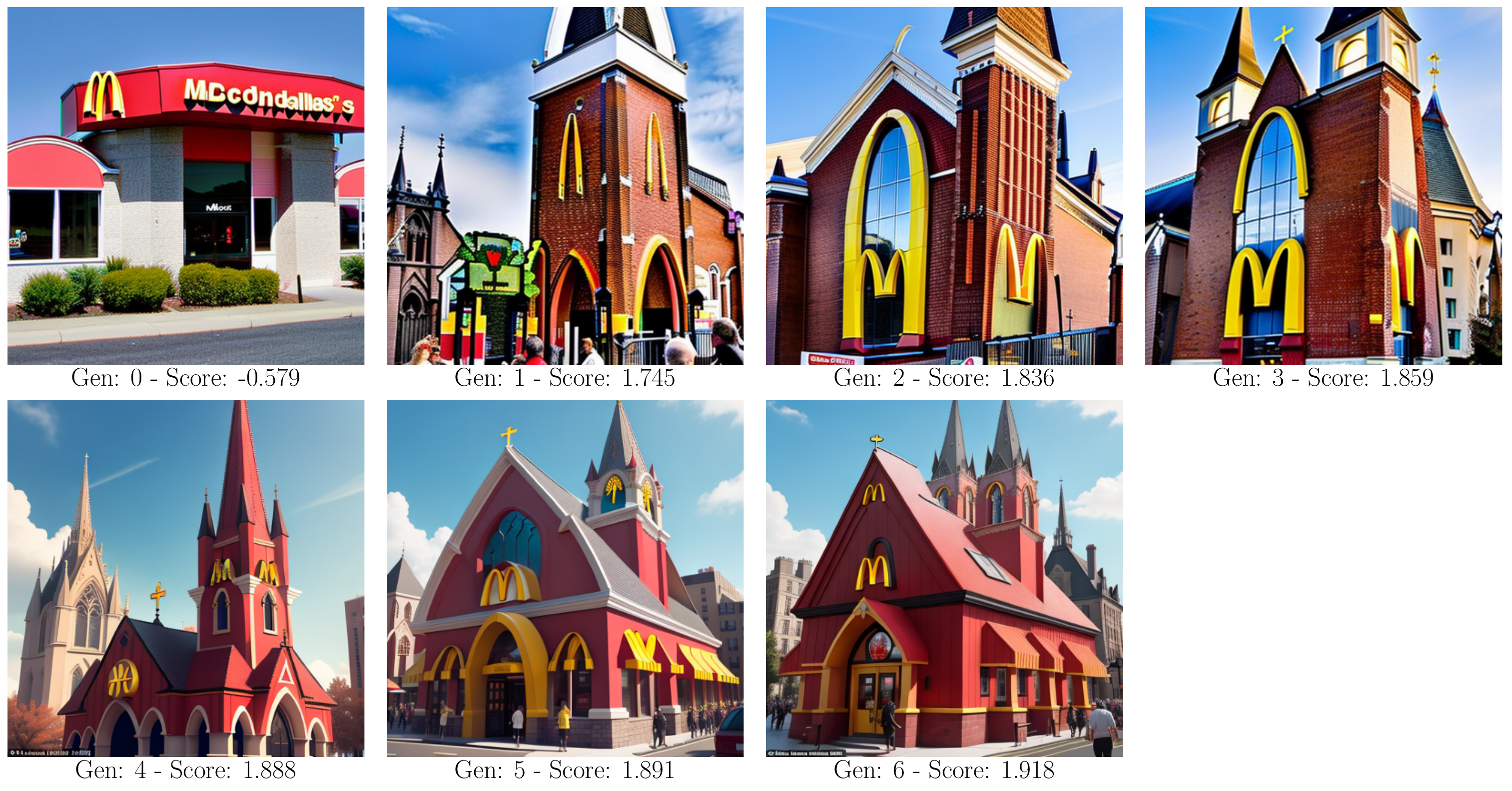}
    \vspace{-8mm} 
    \caption{An example for image improvement with ComfyGI over several generations for the prompt \textit{``mcdonalds church''}.}
    \label{fig:mcdonalds_church_evolution}
\end{figure*}

\begin{figure*}[!h]
    \centering
    \includegraphics[width=\textwidth]{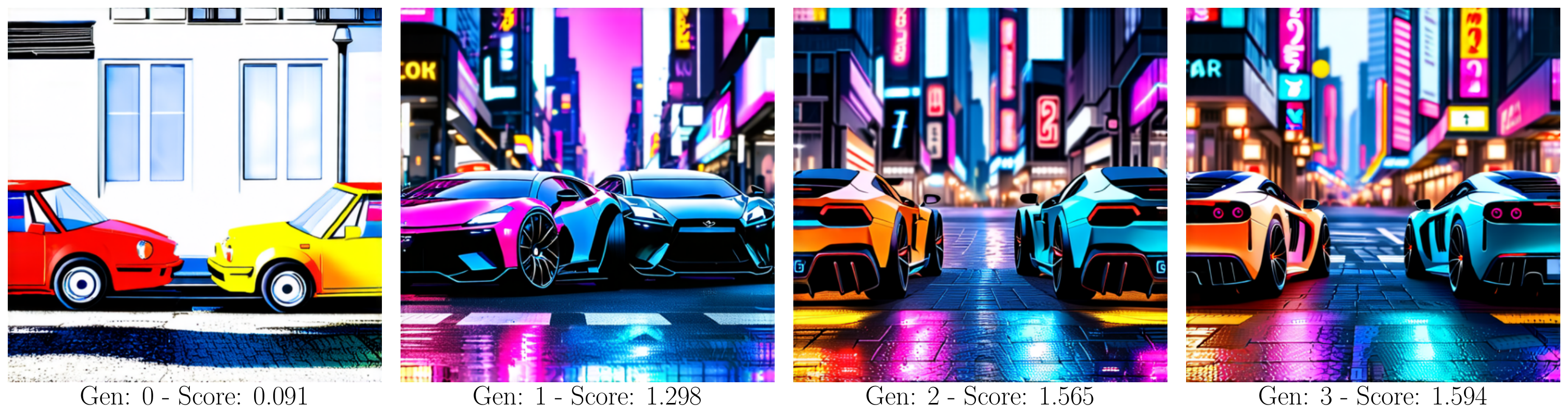}
    \vspace{-8mm} 
    \caption{An example for image improvement with ComfyGI over several generations for the prompt \textit{``two cars on the street''}.}
    \label{fig:two_cars_on_the_street_evolution}
\end{figure*}

Figure~\ref{fig:run_examples} in Sect.~\ref{sec:experiments_results} shows some image examples. However, due to space limitations, we only show the initial and optimized images. Therefore, Figs.~\ref{fig:panda_latte_art_evolution},~\ref{fig:mcdonalds_church_evolution},~and~\ref{fig:two_cars_on_the_street_evolution} show the intermediary steps for these examples.


\clearpage
\section{Additional Plots for the Image Generation Runs}
\label{app:additional_analyses_image_generation}

This section presents analyses that did not fit into the main body of the paper due to space limitations. We show plots that illustrate the performance as well as the model and mutation operator usage at category level. 

\begin{figure*}[!h]
    \centering
    \includegraphics[width=\textwidth]{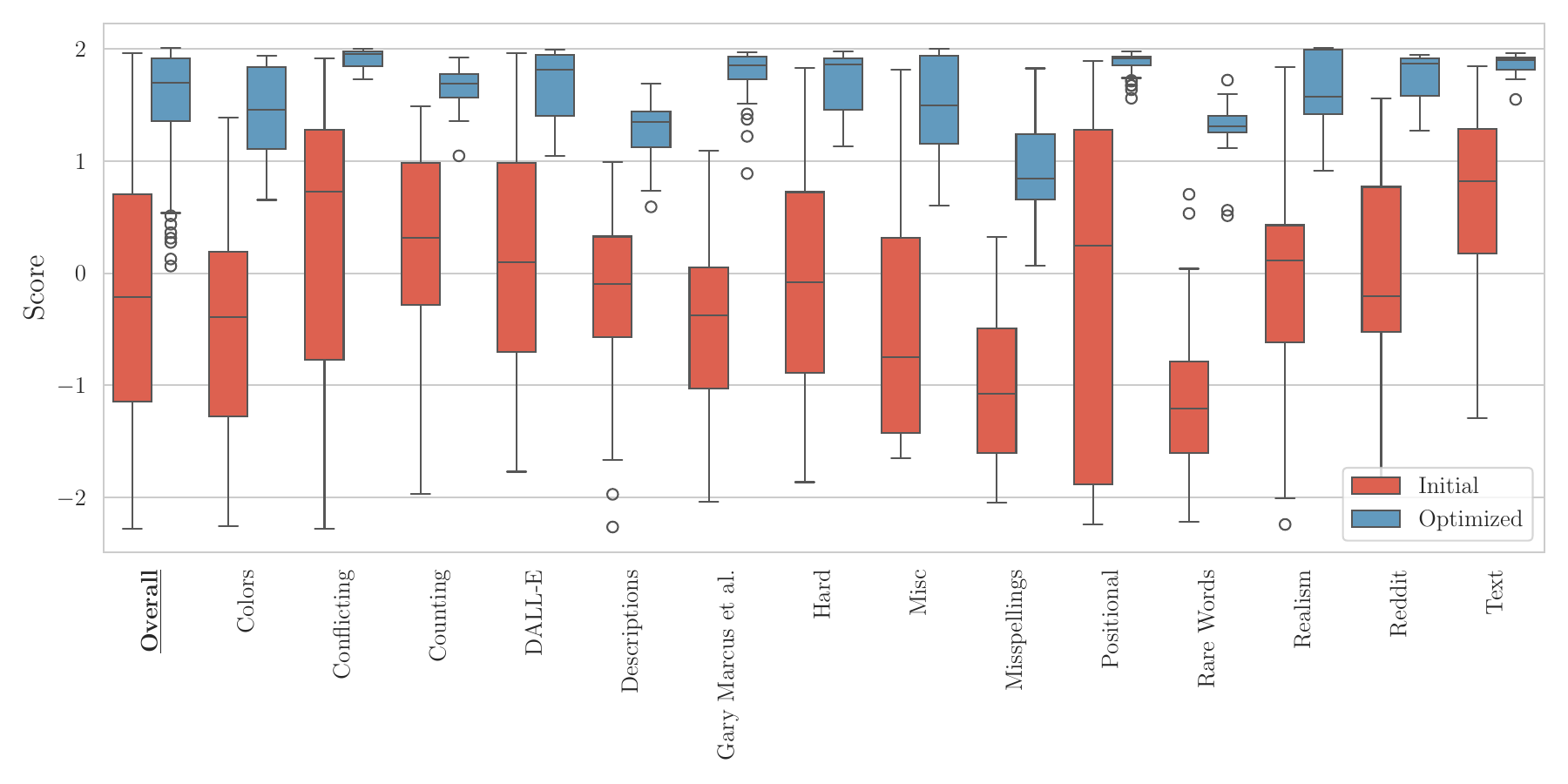}
    \caption{Box-plots of the scores for the initial and optimized images over all prompt categories and runs.}
    \label{fig:score_box_plots_by_category}
\end{figure*}

Figure~\ref{fig:score_box_plots_by_category} shows box-plots of the ImageReward score for the initial and optimized images over all prompt categories and runs. We see, that for each category, the median ImageReward score is better for the optimized images. In addition, also the variance is lower for the optimized images.

\begin{figure*}[!h]
    \centering
    \includegraphics[width=\textwidth]{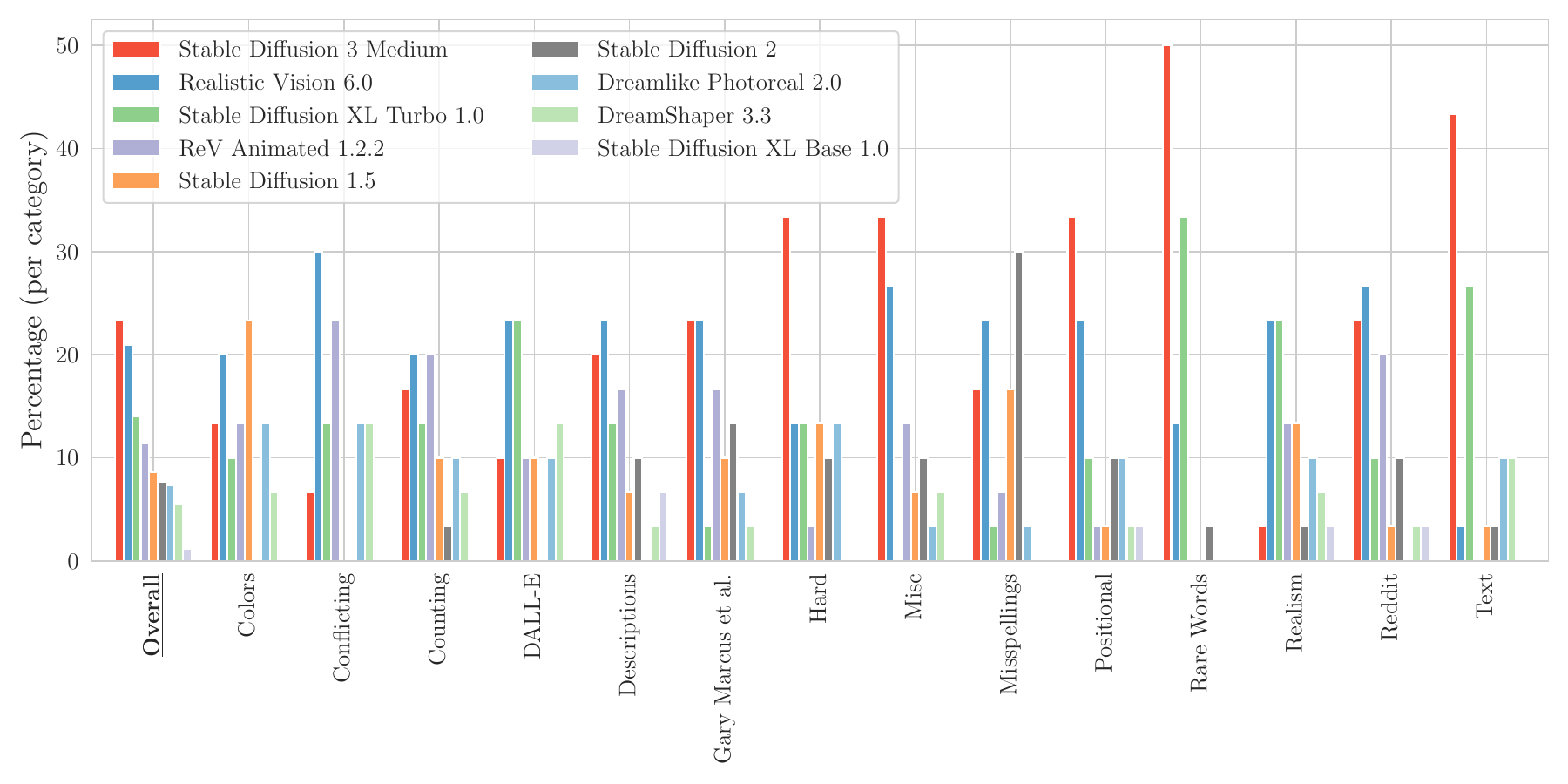}
    \caption{Percentage (per category) of the models used to generate the optimized images over all prompt categories and runs.}
    \label{fig:model_bar_plot_by_category}
\end{figure*}

Figure~\ref{fig:model_bar_plot_by_category} shows the percentage of the models used to generate the optimized images per category. We see that the model \textit{Stable Diffusion 3 Medium} is often selected. E.g., for the categories \textit{Hard}, \textit{Misc}, \textit{Positional}, \textit{Rare Words}, and \textit{Text}. For other categories, different models are selected more often, like \textit{Realistic Vision 6.0}, \textit{Stable Diffusion 2}, or \textit{ReV Animated 1.2.2}. This confirms that ComfyGI is able to select the model that is most suitable to realize the respective target prompt. 

\begin{figure*}[!h]
    \centering
    \includegraphics[width=\textwidth]{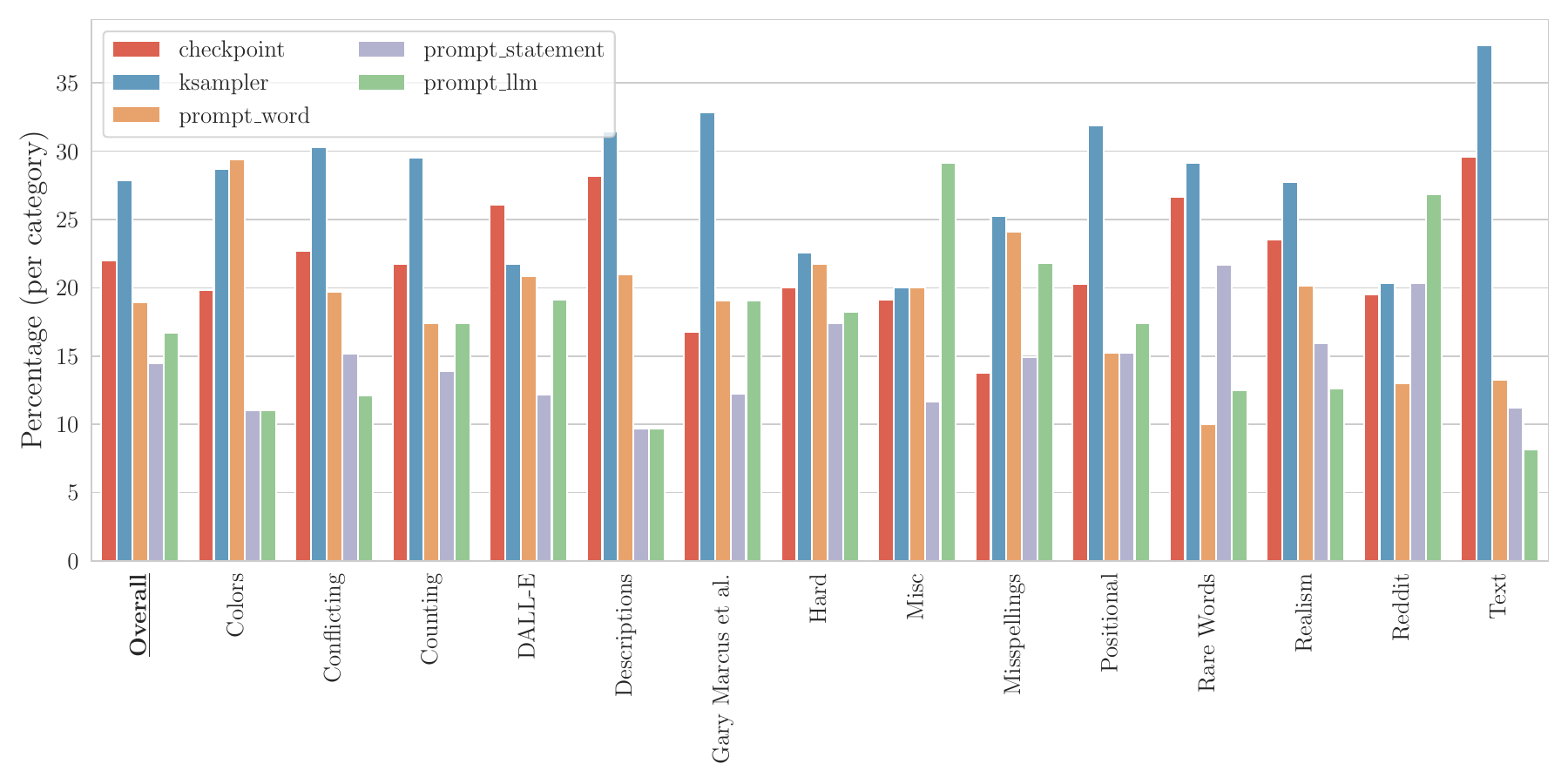}
    \caption{Percentage (per category) of the applied mutations to generate the patches for the workflow optimization over all prompt categories and runs.}
    \label{fig:patch_bar_plot_by_category}
\end{figure*}

Figure~\ref{fig:patch_bar_plot_by_category} shows the percentage of the applied mutations to generate the patches for the workflow optimization per category. We see that overall the \texttt{ksampler} mutation operator is applied most frequently. However, this is not surprising as it makes sense to fine-tune properties like the number of steps or the CFG multiple times in the process of patch generation. Mutation operators like \texttt{prompt\_llm} are applied less frequently although they are very effective. The reason for this is that it is often sufficient to improve the prompt once with an LLM. After that, mutation operators such as \texttt{prompt\_word} or \texttt{prompt\_statement} can further improve the workflow. 

\clearpage
\section{Results of the Human Evaluation}
\label{app:results_human_evaluation}

\begin{figure*}[!ht]
    \centering
    \includegraphics[width=\textwidth]{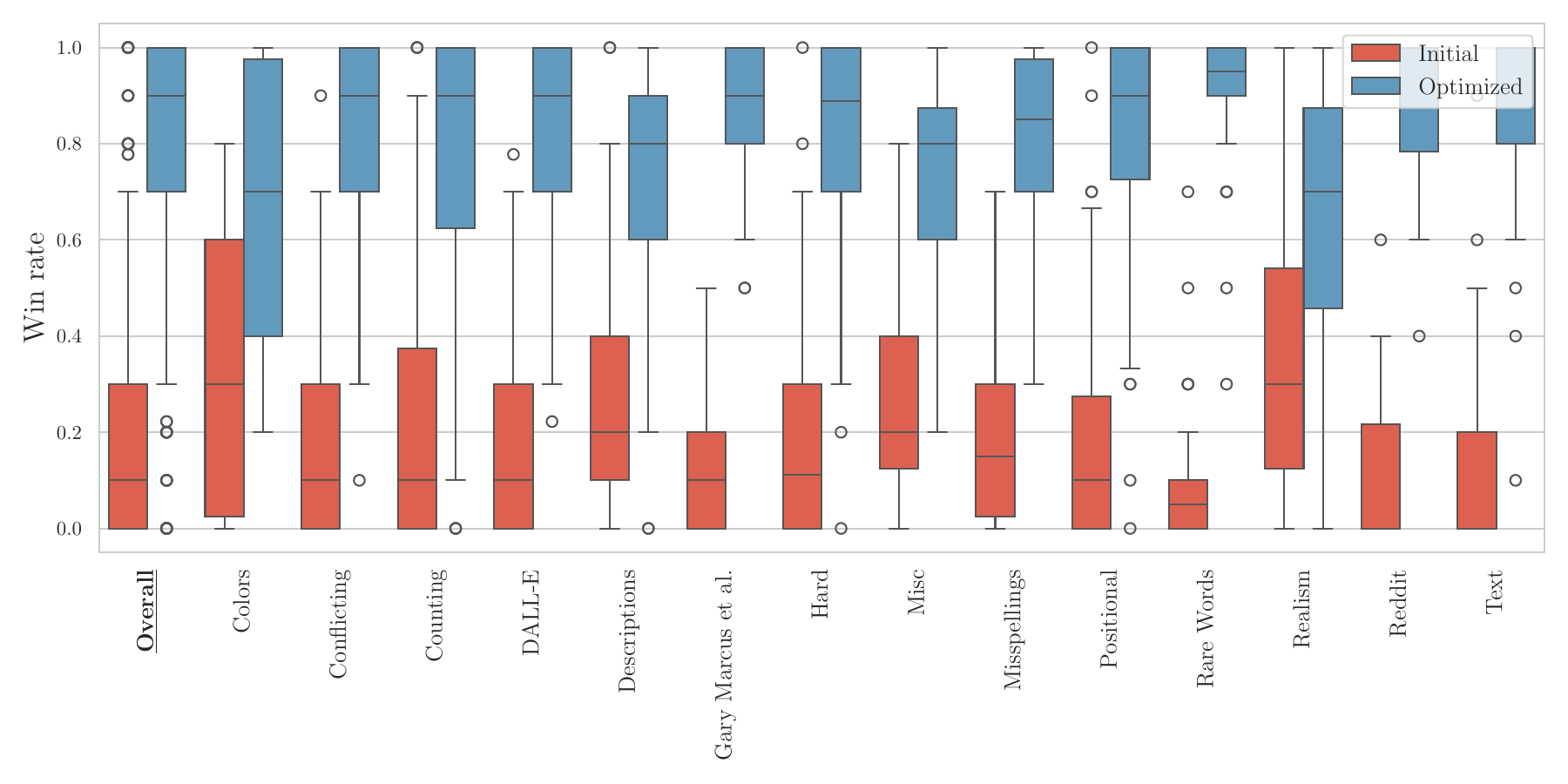}
    \caption{Box-plots of the win rate in the human evaluation for the initial and optimized images over all prompt categories and runs.}
    \label{fig:humaneval_box_plots_by_category}
\end{figure*}

In addition to the analyses with the ImageReward score, we also carried out a human evaluation. Figure~\ref{fig:humaneval_box_plots_by_category} shows box-plots of the win rate for the initial and optimized images per category. As before, we see that the optimized images in each category were also preferred in the human evaluation. 

Table~\ref{tab:descriptives} presents descriptive statistics of the human annotators, like gender, age, education, or employment information.

\onecolumn
\begin{center}
\begin{longtable}{p{15.3cm} c}
\caption{Descriptive statistics of human annotators. \textbf{*English Proficiency} = "How would you rate your English language proficiency from 1 = very poor to 7 = fluent?" \textbf{**Knowledge Text-To-Image AI} = "How would you rate your knowledge regarding text-to-image models such as DALL-E, Midjourney, Stable Diffusion or similar on a scale from 1-7, where 1 is non-existent and 7 is expert-level." \textbf{***Usage Text-To-Image AI} = "How often do you use applications of text-to-image generation models such as DALL-E, Midjourney, Stable Diffusion or similar?"} 
\label{tab:descriptives} \\
\toprule
\endfirsthead

\multicolumn{2}{c}%
{{\tablename\ \thetable{} -- continued from previous page}} \\
\toprule
\endhead

\hline \multicolumn{2}{r}{{Continued on next page}} \\ \hline
\endfoot

\hline \hline
\endlastfoot

    \textbf{N} & 100 \\
        \midrule
    \textbf{Gender} &  \\
    \>Male & 57 \\
    \>Female & 43 \\
    \>Other & 0 \\
    \midrule
    \textbf{Age} & \\
    \>Mean & 40.87 \\
    \>Median & 39 \\
    \>Std.Dev. & 12.498 \\
    \>Min & 18 \\
    \>Max & 74 \\
    \midrule
    \textbf{Education} & \\
    \>Less than high school degree & 0 \\
    \>High school degree or equivalent (e.g. GED) & 29 \\
    \>Some college but no degree & 19 \\
    \>Associate degree & 4 \\
    \>Bachelor degree & 39 \\
    \>Graduate degree & 9 \\
    \midrule
    \textbf{Employment} & \\
    \>Pupil & 0 \\
    \>Student & 4 \\
    \>Apprentice & 0 \\
    \>Employed & 76 \\
    \>Not employed & 11 \\
    \>Retired & 5 \\
    \>Disabled, not able to work & 4 \\
    \midrule
    \textbf{Income} & \\
    \>250.00 \$ or less & 7 \\  
    \>250.01 \$ to 500.00 \$ & 4 \\
    \>500.01 \$ to 750.00 \$ & 5 \\
    \>750.01 \$ to 1,000.00 \$ & 5 \\
    \>1,000.01 \$ to 1,500.00 \$ & 11 \\
    \>1,500.01 \$ to 2,000.00 \$ & 11 \\
    \>2,000.01 \$ to 2,500.00 \$ & 12 \\
    \>2,500.01 \$ to 3,000.00 \$ & 10 \\
    \>3,000.01 \$ to 3,500.00 \$ & 8 \\
    \>3,500.01 \$ to 4,000.00 \$ & 3 \\
    \>4,000.01 \$ to 4,500.00 \$ & 3 \\
    \>4,500.01 \$ to 5,000.00 \$ & 4 \\
    \>5,000.01 \$ or more \$ & 8 \\
    \>Prefer not to tell & 9 \\
    \midrule
    \textbf{Spoken Language} & \\
    \>English & 100 \\
    \\
    \textbf{English Proficiency*} & \\
    \>Mean & 6.95 \\
    \>Median & 7 \\
    \>Std.Dev. & 0.261 \\
    \>Min & 5 \\
    \>Max & 7 \\
    \midrule
    \textbf{Knowledge Text-To-Image AI**} & \\
    \>Mean & 3.39 \\
    \>Median & 4 \\
    \>Std.Dev. & 1.723 \\
    \>Min & 1 \\
    \>Max & 7 \\
    \midrule
    \textbf{Usage Text-To-Image AI***} & \\
    \>Every day & 1 \\
    \>Several times a week & 3 \\
    \>Once a week & 9 \\
    \>Once a month & 20 \\
    \>Less often & 30 \\
    \>Never & 25 \\
    \>I don’t know any of these applications & 12 \\
    \midrule
    \textbf{Attention Checks Passed} & \\
    \>Attention Check 1 & 100 \\
    \>Attention Check 2 & 100 \\
    \>Attention Check 3 & 99 \\
\end{longtable}
\end{center}
\twocolumn

\end{document}